\documentclass{article}

\PassOptionsToPackage{numbers, compress}{natbib}

\usepackage[preprint]{neurips_data_2024}

\usepackage[utf8]{inputenc} % allow utf-8 input
\usepackage[T1]{fontenc}    % use 8-bit T1 fonts
\usepackage{hyperref}       % hyperlinks
\usepackage{url}            % simple URL typesetting
\usepackage{booktabs}       % professional-quality tables
\usepackage{amsfonts}       % blackboard math symbols
\usepackage{nicefrac}       % compact symbols for 1/2, etc.
\usepackage{microtype}      % microtypography
\usepackage{xcolor}         % colors

\usepackage{amsmath}
\usepackage{graphicx}
\usepackage{dsfont}

\usepackage{subcaption}

\title{Seq-to-Final: A Benchmark for Tuning\\from Sequential Distributions\\to a Final Time Point}

\author{Christina X Ji\\MIT CSAIL and IMES\\Cambridge, MA\\\texttt{cji@mit.edu}\\
\And
Ahmed M Alaa\\UC Berkeley and UCSF\\Berkeley, CA\\\texttt{amalaa@berkeley.edu}\\
\And
David Sontag\\MIT CSAIL and IMES\\Cambridge, MA\\\texttt{dsontag@csail.mit.edu}}

\begin{document}

\maketitle

\begin{abstract}

Distribution shift over time occurs in many settings. Leveraging historical data is necessary to learn a model for the last time point when limited data is available in the final period, yet few methods have been developed specifically for this purpose. In this work, we construct a benchmark with different sequences of synthetic shifts to evaluate the effectiveness of 3 classes of methods that 1) learn from all data without adapting to the final period, 2) learn from historical data with no regard to the sequential nature and then adapt to the final period, and 3) leverage the sequential nature of historical data when tailoring a model to the final period. We call this benchmark Seq-to-Final to highlight the focus on using a sequence of time periods to learn a model for the final time point. Our synthetic benchmark allows users to construct sequences with different types of shift and compare different methods. We focus on image classification tasks using CIFAR-10 and CIFAR-100 as the base images for the synthetic sequences. We also evaluate the same methods on the Portraits dataset to explore the relevance to real-world shifts over time. Finally, we create a visualization to contrast the initializations and updates from different methods at the final time step. Our results suggest that, for the sequences in our benchmark, methods that disregard the sequential structure and adapt to the final time point tend to perform well. The approaches we evaluate that leverage the sequential nature do not offer any improvement. We hope that this benchmark will inspire the development of new algorithms that are better at leveraging sequential historical data or a deeper understanding of why methods that disregard the sequential nature are able to perform well.

\end{abstract}

\section{Introduction}
\label{sec:seq_shift_intro}

The state of the world is constantly evolving. For example, fashion and hairstyle trends change over the decades. In healthcare, new diseases and treatments emerge over time. Machine learning models trained on older data may not be able to accurately classify images or forecast major health events in the current time period due to distribution shift over time. To prevent this issue, one option is to train a model only on recent data. However, limited recent data may be insufficient for training a good model. To combat this limitation, models can be pre-trained on other data and then fine-tuned to a particular task. That is how foundation models, such as large language models or text-to-image systems, are trained. For smaller models though, it may be more useful to train the model on data specific to the prediction task. When the dataset is affected by temporal distribution shift, how can we leverage the sequence of historical datasets to train a machine learning model that performs well on the final time point?

Each time period can be considered a single domain with its own data distribution. Many machine learning methods have been developed to leverage data from different domains. Transfer learning is the process of adapting a model pre-trained on a single source domain to a single target domain \citep{pan2010survey}. Many approaches also leverage multiple source domains to learn a model that can generalize to unseen distributions \citep{wang2022generalizing,arjovsky2019invariant,muandet2013domain,rahimian2019distributionally}. When data is presented sequentially, continual learning is often performed to update a model at each time step in a way that continues to perform well on previous domains---a principle known as catastrophic forgetting \citep{kirkpatrick2017overcoming,zenke2017continual,chaudhry2018efficient}.

None of these approaches are designed specifically for leveraging a sequence of historical datasets to perform well at the final time point. A couple benchmarks have been developed to compare the performance of various methods on real-world datasets. The CLEAR benchmark evaluates the performance of many continual learning methods on a set of real-world images collected from Flickr from 2004 to 2014 \citep{lin2021clear}. The Wild-Time benchmark assesses performance at the final time step of many continual learning approaches as well as multi-source transfer learning methods on several real-world datasets over time, including yearbook portraits, satellite images of land usage, Huffpost news articles, arxiv articles, and intensive care unit data \citep{yao2022wild}.

In our benchmark, we focus on image classification tasks and include some of the approaches they evaluate, in addition to other approaches more specialized for adapting to a final distribution. More importantly, to our knowledge, this benchmark is the first to construct building blocks of synthetic shifts that can be put together to form different sequences simulating distribution shift over time. These sequences can then be used to study many types of shifts over time. The benchmark is available at \url{https://github.com/clinicalml/seq_to_final_benchmark}. As a demonstration, we construct 5 four-step synthetic sequences with different types of shifts and 7 synthetic sequences of different lengths with the same type of shift. We also validate that conclusions from our synthetic experiments are relevant by comparing with results on a real-world sequence.

Our benchmark includes 16 methods that 1) learn from all distributions without adapting to the final time point, 2) learn from historical data and then fine-tune on the final time point, and 3) leverage the sequential nature of historical data by sequentially fine-tuning or learning models for all time points jointly. From these experiments, we conclude that current methods that leverage the sequential nature of historical datasets do not provide significant benefits. Oftentimes, learning from all the historical data and then fine-tuning on the final distribution is a good strategy for performing well at the final step. Our work suggests there is a gap in our understanding of how learning from historical data affects the initialization and parameter updates at the final time point.

\section{Benchmark Construction}
\label{sec:benchmark_construction}

To understand how the benchmark is constructed, we first define the sequence-to-final learning problem. Then, we introduce building blocks for different types of shift. Next, we give examples of sequences that can be constructed from these building blocks. Finally, we include a real-world sequence of distribution shifts over time to demonstrate how findings from these synthetic sequences are relevant for real-world applications.

\subsection{Problem Definition}
\label{sec:problem_def}

Suppose we observe a sequence of datasets $\left\{ \mathcal{D}_t \right\}_{t=1}^T = \left\{ \left(x_t^{\left(i\right)}, Y_t^{\left(i\right)}\right)_{i=1}^{n_t} \overset{iid}{\sim} \mathbb{P}_t\left(X, Y\right) \right\}_{t=1}^T$. We assume all samples from a time period $t$ come from the same distribution $\mathbb{P}_t$, that is, there is no distribution shift within a time period. This means there is no gradual shift, and the time periods are divided where shifts occur. For our benchmark, $x_t^{\left(i\right)}$ is an image, and $Y_t^{\left(i\right)}$ is a discrete label. The goal is to learn a model that minimizes the loss at the final time point:
\begin{equation}
    \label{eq:final_objective}
    \min_{f \in \mathcal{F}} \mathcal{L}\left(f, \mathcal{D}_T\right)
\end{equation}
Unlike in continual learning, the goal is not to prevent catastrophic forgetting, so historical datasets do not appear in the objective. However, historical datasets can help with learning $f$ because we assume samples from closer time steps have more similar distributions with some leeway from $\epsilon$:
\begin{equation}
   \exists \epsilon > 0, \, d\left(\mathbb{P}_{t-j}, \mathbb{P}_t\right) < d\left(\mathbb{P}_{t-k}, \mathbb{P}_t\right) + \epsilon \,\, \forall \, 1 \le j < k \label{eq:shift_increases_dist}
\end{equation}
where $d: \mathbb{P} \times \mathbb{P} \rightarrow \mathbb{R}$ is a distance metric between distributions, such as the Wasserstein-2 distance we use to quantify shift in Appendix~\ref{app:quantifying_shift}. This assumption means changes are not reversed without other changes being introduced.

\subsection{Building Blocks: Types of Shift}
\label{sec:shift_building_blocks}

To construct sequences of synthetic shifts, we first need to consider the types of shifts that may occur in practice and how they affect which parts of the model need to be adapted. Inspired by \citet{lee2022surgical}, we create 6 building blocks that fall under 3 classes of shift:

\textbf{Input-level shift} We consider three types of input-level shifts: 1) \textit{Corruptions (C)}: We randomly draw discrete noise between -3 and +2 inclusive for each pixel in each channel and add it to the original pixel value. 2) \textit{Rotations (R)}: Each image is rotated 30 degrees counterclockwise. 3) \textit{Red tinting (T)}: We add 30 to every pixel in the red channel. The green and blue channels are unaffected. Figure~\ref{fig:crt_repeat_examples} in Appendix~\ref{app:image_examples} illustrates these transformations. These changes can be considered covariate shifts as the change is in $\mathbb{P}\left(X\right)$. They can usually be addressed by changing model layers closer to the input so the transformed features can map to the original representation \citep{lee2022surgical}.

\textbf{Output-level shift} We include one type of output-level shift: 1) \textit{Label flips (L)}: We flip the label classes in the same way as \citet{lee2022surgical}. For instance, if there are 10 classes, the labels are changed to $9 - Y$. For additional steps with label flips, our code alternates between changing the labels to $9 - Y$ and $Y + 2$ (with 8 changing to 0 and 9 changing to 1) so the labels do not revert to the previous step. Note that this is not label shift as $\mathbb{P}\left(Y\right)$ stays the same when all label classes have the same frequency. These changes can usually be addressed by only changing the layer closest to the output \citep{lee2022surgical}. The feature representation does not need to change. 

\textbf{Intermediate-level shift} We introduce two types of intermediate-level shifts: 1) \textit{Conditional rotations (r)}: The images are rotated 30 degrees clockwise for some labels and 30 degrees counterclockwise for other labels. 2) \textit{Sub-population shifts (s)}: Within each label class, there may be different sub-populations. A sub-population shift either introduces a new type of that label or replaces one type of that label with another type. This corresponds to feature-level shifts in \citet{lee2022surgical}. Because both of these transformations depend on the label class, they are conditional shifts---that is, changes in $\mathbb{P}\left(X \vert Y\right)$. We denote these shifts with lower case r and s to differentiate the conditional aspect. Addressing these shifts requires larger changes in the model.

\subsection{Sequences of Synthetic Shifts}
\label{sec:sft_synthetic_setup}

To construct a sequence of shifts, we can choose different building blocks to put together. To illustrate, we create 5 sequences each with 4 time steps: 1) \textit{Corruption, label flip, rotation (CLR)}: This sequence combines different types of shifts, so the model will need to adapt different layers at each time step. 2) \textit{Rotation, corruption, label flipping (RCL)}: The images at the final time step are the same as those in the previous sequence. However, applying label flips last means different adaptations need to be made at the final step. 3) \textit{Conditional rotation (rrr)}: This sequence repeatedly rotates images 30 degrees, with the direction conditional on the label. 4) \textit{Red tinting (TTT)}: This sequence repeatedly adds 30 to the red channel. 5) \textit{Sub-population shifts (sss)}: Step 0 starts with 2 sub-classes for each label class. Step 1 introduces a new sub-class for each label. Steps 2 and 3 each replace an original sub-class with a new sub-class. The 3 sub-classes at the final step are distinct from the 2 initial sub-classes. Appendix~\ref{app:image_examples} visualizes examples from these sequences, and Appendix~\ref{app:quantifying_shift} quantifies the amount of covariate and conditional shift over time.

The first four sequences are built on CIFAR-10, a collection of 32 x 32 color images from 10 label classes \citep{krizhevsky2009cifar}. The last sequence is built on CIFAR-100 to utilize sub-class labels \citep{krizhevsky2009cifar}. CIFAR-100 is a collection of 32 x 32 color images from 20 label super-classes, each of which is divided into 5 fine label classes. We select these datasets because they are commonly used in the machine learning field. We choose to create sequences with 6000 samples in the source distribution, 4000 samples at step 1, 6000 samples at step 2, and 4000 samples at step 3. 4000 samples at the final step is insufficient to learn a good model, so leveraging historical data is critical. At each step, the images are randomly drawn from the training samples, and 20\% of the samples are placed in a validation set for tuning hyper-parameters. As defined in Equation~\ref{eq:final_objective}, the objective is to perform well at the final time point. Therefore, we draw 5000 samples from the test images to evaluate the test accuracy  at the final step.

To study the effect of sequence length, we additionally construct sequences of \textit{two to eight 30-degree rotations}. For a fair comparison, the sample size at the final step is fixed at 4000. The remaining 16000 samples are divided among the other steps as listed in Table~\ref{tab:rotation_seq_sample_sizes} in Appendix~\ref{app:image_examples}.

\subsection{Sequence of Real-World Shifts}
\label{sec:real_world_seq}

To examine whether conclusions from synthetic sequences constructed in our benchmark apply to real-world shifts, we evaluate the same methods on a real-world sequence built on the \textit{Portraits} dataset \citep{ginosar2015century}. This dataset has been used to study temporal distribution shift by \citet{kumar2020understanding} and \citet{yao2022wild} since it spans 1930 to 2013. The task is to classify yearbook pictures by gender. We make a couple modifications to the set-up in the Wild-Time benchmark \citep{yao2022wild}: 1) We define a single time period to be 1 decade instead of 5 years so there are 8 time periods each with 2500 to 5500 samples. The exact sample sizes are shown in Table~\ref{tab:quantify_shifts_real} in Appendix~\ref{app:quantifying_shift}. The final time period includes samples from 2000-2013. This way, the sequence length and sample sizes are more similar to the synthetic sequences. Figure~\ref{fig:portraits_visual} in Appendix~\ref{app:image_examples} shows examples from each time period. 2) At earlier time steps, a test split is not needed. The samples are split 80/20 between training and validation. At the final time period, 50\% of samples are held out for the test set.

\section{Methods to Learn from Data with Sequential Distribution Shift}

\label{sec:methods}

In our benchmark, we compare three classes of methods that leverage historical data: 1) methods that learn from all data with no adaptations for the final step, 2) methods that learn from all historical data and then adapt to the final step, and 3) methods that leverage the sequential nature of historical data and target the final step. All the methods are summarized in Table~\ref{tab:methods}, and additional details are provided in Appendix~\ref{app:benchmark_methods}.

\begin{table}[htbp]
    \centering
    {\footnotesize 
    \begin{tabular}{lp{4cm}cp{2.78cm}p{2.78cm}}
    \toprule
    Method & Description & Class & Historical mode & Final mode \\
    \midrule
    Oracle & ERM on 20k final samples & Oracle & Not used & ERM \\
    Baseline & ERM on 4k final samples & Baseline & Not used & ERM \\
    \midrule
    ERM & Empirical risk minimization & 1 & \multicolumn{2}{c}{---------- ERM on all steps ----------} \\
    IRM & Invariant risk minimization & 1 & \multicolumn{2}{c}{---------- IRM on all steps ----------} \\
    DRO & Distributionally robust optimization & 1 & \multicolumn{2}{c}{---------- DRO on all steps ----------} \\
    \midrule
    FT & ERM + fine-tune on final & 2 & ERM & Fine-tune \\
    LP-FT & ERM + linear probe then fine-tune on final & 2 & ERM & Linear probe then fine-tune \\
    I-FT & IRM + fine-tune on final & 2 & IRM & Fine-tune \\
    D-FT & DRO + fine-tune on final & 2 & DRO & Fine-tune \\
    ST-1 & ERM + side-tune 1-layer on final & 2 & ERM & Fit 1-layer side modules \\
    ST-B & ERM + side-tune block on final & 2 & ERM & Fit side blocks \\
    \midrule
    SFT & Sequential fine-tuning & 3 & ERM on source, then fine-tune each step & Fine-tune \\
    EWC & Elastic weight consolidation & 3 & ERM on source, then fine-tune each step, regularize to previous & Fine-tune, regularize to previous \\
    SST-1 & Sequential side-tuning 1-layer & 3 & ERM on source, then add 1-layer side modules at each step & Add 1-layer side modules \\
    SST-B & Sequential side-tuning block & 3 & ERM on source, then add side blocks at each step & Add side blocks \\
    \midrule
    JM & Joint model & 3 & \multicolumn{2}{p{5.56cm}}{Separate modules for each step with adjacent steps regularized to be closer} \\
    JST-1 & Joint model with 1-layer side modules & 3 & \multicolumn{2}{p{5.56cm}}{Original modules at step 0 and additive 1-layer side modules for each subsequent step} \\
    JST-B & Joint model with side blocks & 3 & \multicolumn{2}{p{5.56cm}}{Original modules at step 0 and additive blocks for each subsequent step} \\
    \bottomrule
    \end{tabular}}
    \caption{Description of methods in benchmark}
    \label{tab:methods}
    \vspace{-20pt}
\end{table}

\subsection{Oracle and Baseline}
\label{sec:oracle_baseline}

To provide upper and lower bounds on the test accuracy that can be achieved, we include methods that only use data from the final step:

\textbf{Oracle} For a sequence with a total of $n$ samples across all steps, the oracle has access to $n$ samples all from the final step. The oracle runs empirical risk minimization (ERM) on all of these samples. Methods that are better at leveraging historical data would have performance closer to this oracle.

\textbf{Baseline} The baseline discards historical data and runs ERM on the limited data at the final step.

\subsection{Learning from All Distributions}
\label{sec:methods_from_all_dist}

This first class of methods utilizes data from multiple time steps and does not pay special attention to the final time step. We include a couple approaches that train specifically for domain generalization, where the goal is to leverage multiple source domains to generalize to an unseen distribution.

\textbf{Empirical risk minimization on all steps (ERM)} This method groups data from the final step with all historical data and performs ERM on this combined dataset without targeting the final step.

\textbf{Invariant risk minimization on all steps (IRM)} The goal of IRM is to learn a model that captures features that are constant across all domains \citep{arjovsky2019invariant}. We apply IRM to all historical time points and the final step to learn features that may persist at the final step.

\textbf{Distributionally robust optimization on all steps (DRO)} This method learns a model that minimizes the worst loss across all time steps, whether that is a historical time point or the final step.

\subsection{Pre-training with Historical Data and Adapting to the Final Distribution}
\label{sec:methods_adapt_final}

The second class of methods learns from historical data and then adapts to the final distribution. Model weights are initialized at the parameters learned from historical data and then fine-tuned on samples from the final step. To improve data efficiency and avoid unnecessary updates, some methods only update particular layers or add a small set of parameters that is learned for the new distribution while the original parameters are kept frozen \citep{lee2022surgical,von2021learning,hu2021lora,liu2022few,zhang2020side}. Typically, the model is initialized by performing ERM on historical data. Because there are multiple historical steps, we also consider initializations from running IRM and DRO on the historical data.

\textbf{Fine-tuning (FT)} This method starts with ERM on all historical data and then updates all the parameters by performing gradient descent on samples from the final step. This allows for the greatest flexibility but may result in over-fitting to the limited target data.

\textbf{Linear probing then fine-tuning (LP-FT)} This method starts with ERM on all historical data. Then, as \citet{kumar2022fine} proposed, only the layer closest to the output is tuned. After that reaches convergence, all the layers are fine-tuned. \citet{kumar2022fine} suggest that linear probing first may help prevent feature distortion. \citet{lee2022surgical} show that linear probing may be sufficient for output-level shifts, such as the label flipping shift in our benchmark.

\textbf{IRM then fine-tuning (I-FT)} This method starts by learning an invariant feature representation with IRM on the historical domains. Then, it fine-tunes this pre-trained model on the final step. To our knowledge, domain generalization approaches have not been used as initializations for fine-tuning. We hypothesize these initializations may be better since they can capture patterns unaffected by temporal shift.

\textbf{DRO then fine-tuning (D-FT)} This method starts by minimizing the worst-case loss on all historical steps with DRO. Then, it fine-tunes on the final step. Similar to IRM then fine-tuning, we hypothesize that DRO then fine-tuning will also find a better initialization.

\textbf{Side-tuning with 1-layer side modules (ST-1)} This method fits a model on all the historical data via ERM. Then, for the final step, a side module $S$ is added parallel to each original block $B$, so the new output from each block is $B\left(x\right) + S\left(x\right)$. Each side module $S$ has 1 convolution layer followed by batch normalization. $B$ is frozen at the final step, and $S$ is fit via gradient descent. This is more parameter-efficient than fine-tuning the entire model. We borrow the name ``side-tuning'' from \citet{zhang2020side}. They add a single side module for the entire model, whereas we add a side module for each block. Side modules are also sometimes called parallel adapters \citep{he2021towards,rebuffi2018efficient}.

\textbf{Side-tuning with side blocks (ST-B)} We include a variant where side modules match the original blocks to test whether side-tuning the same number of parameters is comparable to fine-tuning.

\subsection{Leveraging Sequential Historical Data and Targeting the Final Distribution}
\label{sec:methods_leverage_seq}

The third class of methods leverages the sequential nature of the historical datasets. There are two sub-classes. The first sub-class updates the model sequentially at each step. We also introduce parameter-efficient variants in this sequential setting. 

\textbf{Sequential fine-tuning (SFT)} This method starts by fitting a model on the source data. At each subsequent step, the model from the previous step serves as the initialization for fine-tuning. We try different learning rate schedules over the time steps as described in Appendix~\ref{app:benchmark_methods}. 

\textbf{Elastic weight consolidation (EWC)} This continual learning method also sequentially fine-tunes the model at each step. The regularization term encourages parameters that were important for previous tasks to have smaller changes at subsequent time steps to prevent catastrophic forgetting \citep{kirkpatrick2017overcoming}.

\textbf{Sequential side-tuning with 1-layer side modules (SST-1)} This method adds a new side module $S_t$ at each time step $t$. The output from a block at time $t$ is $B\left(x\right) + \sum_{j=1}^t S_j\left(x\right)$. The base module $B$ and the previous side modules $S_1, \ldots, S_{t-1}$ are frozen. $S_t$ is initialized randomly and learned from data at time $t$. To our knowledge, adding side modules repeatedly in this fashion is a novel method.

\textbf{Sequential side-tuning with side blocks (SST-B)} Similar to the motivation for ST-B, we also sequentially add block-sized side modules to evaluate whether sequential side-tuning is comparable to sequential fine-tuning when the number of parameters is the same. 

The second sub-class learns from all distributions at once while enforcing a sequential structure in the models over time. We draw from multi-task learning to model the time steps jointly \citep{zhang2018overview}. Instead of a shared feature representation and different prediction heads for each task, all layers are separate across time steps to handle different types of shifts. The tasks are related by regularization or the side module structure instead. To target the final step, the training objective gives higher weight to the loss at the final step. More details about tuning these joint models are provided in Appendix~\ref{app:benchmark_methods}.

\textbf{Joint model (JM)} To enforce the sequential relationship, this method applies L2 regularization on the difference between parameters at adjacent time steps.

\textbf{Joint model with 1-layer side modules (JST-1)} This joint model has standard blocks at the first step and adds 1-layer side modules $S_t$ at each subsequent step $t$. The output from a block at time $t$ is $B\left(x\right) + \sum_{j=1}^t S_j\left(x\right)$. This joint model has the same structure as the model at the final step of SST-1, but the entire model is learned from all datasets at once.

\textbf{Joint model with side blocks (JST-B)} This joint model adds side blocks $S_t$ at each step. It has the same number of parameters as JM and the same structure as the final model in SST-B. Similar to ST-B and SST-B, this method is included to disentangle the limitations of the additive side module structure from the reduced number of parameters in SST-1.

\begin{table}[tbp]
    \centering
    \begin{tabular}{lccccccc}
    \toprule
    Sequence & Oracle & Baseline & ERM & IRM & DRO \\
    \midrule
    Corruption, label, rotation & .66 & .54 & .51 & .49 & .46 \\
    Rotation, corruption, label & .66 & .54 & .19 & .24 & .30 \\
    Conditional rotation x3 & .72 & .59 & .57 & .57 & .55 \\
    Red tinting x3 & .68 & .55 & .65 & .62 & .59 \\
    Sub-population shift x3 & .55 & .40 & .48 & .48 & .48 \\
    Portraits &  --- & .82 & .86 & \textbf{.93} & .89 \\
    \midrule
    Sequence & FT & LP-FT & I-FT & D-FT & ST-1 & ST-B \\
    \midrule
    Corruption, label, rotation & \textbf{.58} & \textbf{.58} & \textbf{.58} & .57 & .55 & .55 \\
    Rotation, corruption, label & .61 & \textit{\textbf{.64}} & .59 & .57 & \textbf{.63} & .61 \\
    Conditional rotation x3 & \textbf{.63} & \textbf{.63} & \textbf{.63} & \textbf{.64} & \textbf{.62} & \textbf{.63} \\
    Red tinting x3 & \textit{\textbf{.67}} & \textit{\textbf{.67}} & \textit{\textbf{.68}} & \textit{\textbf{.66}} & \textit{\textbf{.68}} & \textit{\textbf{.66}} 
    \\
    Sub-population shift x3 & .51 & .51 & .50 & .51 & \textit{\textbf{.55}} & .50 \\
    Portraits & .89 & \textbf{.93} & \textbf{.92} & .91 & .88 & .90 \\
    \midrule
    Sequence & SFT & EWC & SST-1 & SST-B & JM & JST-1 & JST-B \\
    \midrule
    Corruption, label, rotation & \textbf{.59} & \textbf{.58} & .57 & .57 & .57 & \textbf{.60} & .57 \\
    Rotation, corruption, label & .61 & .59 & .59 & .59 & .58 & \textbf{.63} & .60 \\
    Conditional rotation x3 & \textbf{.63} & \textbf{.62} & .61 & \textbf{.62} & \textbf{.62} & \textbf{.62} & .60 \\
    Red tinting x3 & \textit{\textbf{.66}} & .64 & .64 & .63 & .65 & .65 & .61 \\
    Sub-population shift x3 & \textit{\textbf{.53}} & .52 & \textit{\textbf{.54}} & .50 & .47 & .51 & .48 \\
    Portraits & \textbf{.94} & \textbf{.93} & .88 & .87 & .65 & .66 & .68 \\
    \bottomrule
    \end{tabular}
    \caption[Test accuracy at final step of each sequence.]{Test accuracy at final step of each sequence. Mean is taken over the 12 model architectures in Table~\ref{tab:model_architectures}. Mean is italicized if it reflects oracle-like performance (within .02 or better than the oracle). Mean is bolded if it is among the best-performing methods (oracle-like or within .02 of the best method). The threshold .02 was chosen based on the size of the standard deviations across the 12 architectures. See Tables~\ref{tab:clr_results_full}-\ref{tab:portraits_results_full} in Appendix~\ref{app:full_results} for results from individual architectures. Top: Methods that do not adapt to the final step. Middle: Methods that learn from historical data and adapt to the final step. Bottom: Methods that leverage the sequential nature and target the final step.}
    \label{tab:mean_results}
    \vspace{-20pt}
\end{table}
    
\section{Results}
\label{sec:results}

We evaluate all of the methods described in Section~\ref{sec:methods} on the synthetic and real-world sequences described in Section~\ref{sec:benchmark_construction}. The results for the fixed-length sequences of synthetic shifts and the real-world sequence are shown in Table~\ref{tab:mean_results}. Each class of methods is shown in a different subsection of the table to make it easy to distinguish the best-performing classes for each sequence. These are the key take-aways from these experiments:

\textbf{Leveraging sequential nature may be unnecessary} Learning from all historical data ignoring the sequential structure provides just as good of an initialization for fine-tuning on the final distribution as sequentially updating the model over time. This observation holds in general for all  shift sequences we constructed as there are bolded metrics in every row of the middle section.

\textbf{LP-FT is clear winner when final shift is label flipping} This aligns with the findings in \citet{kumar2022fine} and \citet{lee2022surgical}. Small side modules  may also be effective at preventing feature distortion.
    
\textbf{Joint model does not seem to benefit from accessing all steps at once} The additive side module structure or the regularization bringing weights closer at adjacent time steps may be too constraining.

\textbf{Oracle-level performance can be achieved for sequences with less shift} For the sequences of red tinting and sub-population shifts, many methods achieve oracle-level performance by leveraging historical data. The former sequence has no change in the blue and green channels, and the latter sequence does not appear to have much shift according to the metrics we compute in Appendix~\ref{app:quantifying_shift}.

\textbf{Conclusions from synthetic sequences are relevant to real-world applications} For the real-world Portraits sequence, some methods that do not leverage the sequential nature of historical data are also among the best-performing methods.

More fine-grained interpretations, as well as results for the variable-length synthetic sequences, are in Appendix~\ref{app:full_results}.

\begin{figure}[tbp]
    \centering
    \begin{minipage}{.37\textwidth}
        \begin{subfigure}{\linewidth}
            \centering
            \includegraphics[width=\textwidth,trim={30 320 310 50},clip]{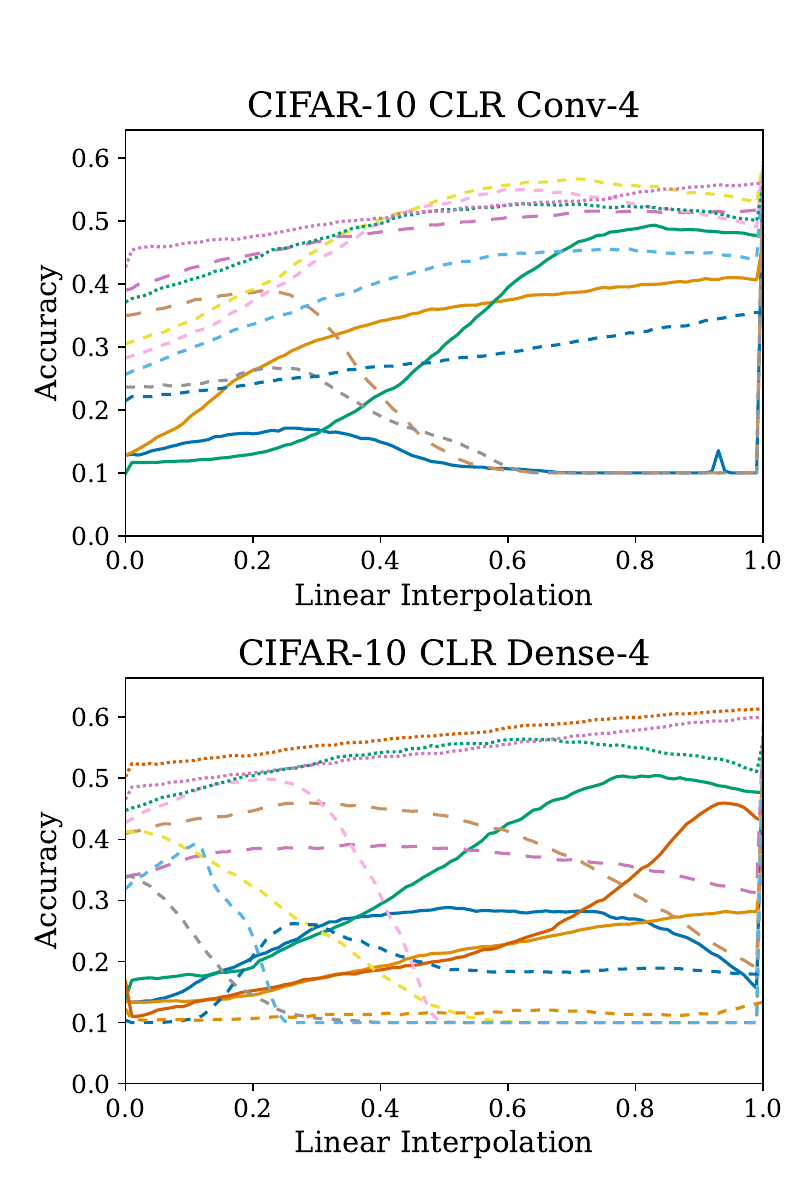}
        \end{subfigure}
    \end{minipage}
    \centering
    \begin{minipage}{.37\textwidth}
        \begin{subfigure}{\linewidth}
            \centering
            \includegraphics[width=\textwidth,trim={30 590 310 45},clip]{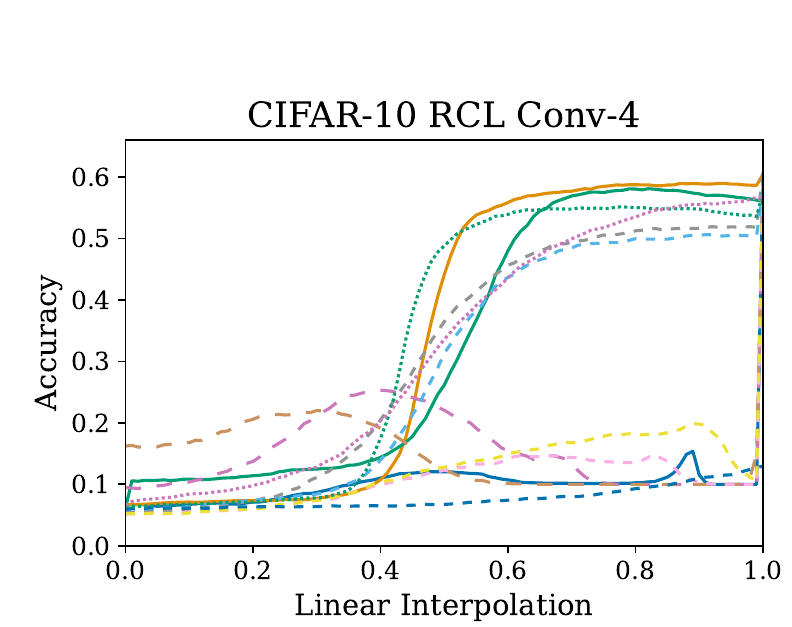}
        \end{subfigure}
        \begin{subfigure}{\linewidth}
            \centering
            \includegraphics[width=\textwidth,trim={30 50 310 610},clip]{figures/addressing_seq_shift_figures/CLR_all_architectures_interpolated_accuracies.pdf}
        \end{subfigure}
    \end{minipage}
    \centering
    \begin{minipage}{.24\textwidth}
        \begin{subfigure}{\linewidth}
            \centering
            \includegraphics[width=\textwidth,trim={410 200 50 200},clip]{figures/addressing_seq_shift_figures/RCL_all_architectures_interpolated_accuracies.pdf}
        \end{subfigure}
    \end{minipage}
    \caption{Linear interpolation paths from initialization to final model at last step.  Top right: Rotation, corruption, and label flip sequence. Other plots: Corruption, label flip, and rotation sequence. Top: 4-block convolutional networks. Bottom left: 4-block dense network. Bottom right: 4-block residual network. Architectures are described in Appendix~\ref{app:image_class_models}.}
    \label{fig:clr_lin_interpolate}
    \vspace{-20pt}
\end{figure}

\subsection{Visual Exploration of Models}

\label{sec:lin_interpolate_visual}

As shown in Table~\ref{tab:methods}, the methods in our benchmark vary in two aspects: 1) the historical mode that determines the initialization at the final step and 2) the updates performed at the final step. We hypothesize that some methods may find an initialization in a loss basin that has a good local optimum for the final step and stay in that basin, while other methods may start at a less ideal initialization and need to cross a performance barrier to enter a better loss basin.

To test this hypothesis, we visualize the loss landscape by linearly interpolating model weights from the initialization to the final model at the last time step and plotting the test accuracy at the last step for each interpolated model. We justify this approach and discuss how \citet{neyshabur2020being} and \citet{mehta2023empirical} use similar visualizations in Appendix~\ref{app:lin_interpolation_visual}.  For side-tuning, we interpolate only the side module weights from 0 to their final values since zero-ing out the side modules matches the model from the previous time point. We call the trajectory of test accuracies as the weights are linearly interpolated the linear interpolation path. These paths can be interpreted to test our hypothesis:

\textbf{Good initialization with little tuning} The path would start close to the optimum and be fairly flat.

\textbf{Change of loss basin} The path would decrease and then increase. (Since higher test accuracy is better, the path would appear to traverse between two plateaus rather than two basins.) The initialization may not be near a good local optimum, or historical information may not be preserved during updates.
    
\textbf{Sharp local minimum} The path would increase sharply when approaching the final weight. This has the same implications mentioned above for the initialization or preservation of historical information.

With these potential interpretations in mind, we now turn to the visualizations in Figure~\ref{fig:clr_lin_interpolate} for two synthetic sequences. Each panel presents the linear interpolation paths for a different model architecture. Visualizations for the other synthetic sequences are shown in Appendix~\ref{app:lin_interpolation_visual}. We can draw the following conclusions from these visualizations:

\textbf{Smooth sigmoid shape for best-performing methods when final step is label flip} There is a clear inflection point where the label predictions flip and the test accuracy increases smoothly.

\textbf{Good initialization followed by steady increase for joint model variants} The regularization that encourages similar weights at adjacent steps helps with putting models at the last two steps in the same loss basin. For joint models with side modules, the additive output structure and back-propagation of losses at the final step through all modules from previous steps may contribute to this smoothness.

\textbf{High variability across different model architectures for some methods} An example is SFT with no decay or linear decay in learning rates for the corruption, label flip, and rotation sequence in Figure~\ref{fig:clr_lin_interpolate}. The paths improve steadily for convolutional networks, cross a performance barrier with 0.1 accuracy for dense networks, and stay flat before increasing sharply at the end for residual networks. Similar variation across architectures is reflected in the test accuracies in Tables~\ref{tab:clr_results_full}-\ref{tab:sss_results_full}. This variability may also be due to random differences across individual runs.

\section{Discussion}

In this work, we introduce a benchmark for evaluating how well different algorithms can leverage historical data to learn a good model for the final time point when the sequence of datasets is affected by distribution shift over time. Our benchmark can be used to construct synthetic sequences of image datasets with different types of shift. We construct several synthetic sequences using our benchmark and evaluate a panel of methods that 1) learn to generalize across all time points without targeting the final step, 2) learn from all historical data and then adapt to the final step, and 3) learn from a sequence of datasets while targeting the final distribution. Among this panel, we find that the second class of methods that does not account for the sequential nature of historical data performs just as well as the third class that leverages the sequential aspect.

We hope that this benchmark can serve as an easy-to-use, reproducible, and modular environment for assessing the performance of new methods to address temporal distribution shift and identifying the sequences of shifts that new methods are better at addressing.

\textbf{Ease of use} We provide command-line arguments for constructing the shift sequence and selecting the method to run.

\textbf{Reproducibility} Every time the same sequence of shifts is constructed with the same random seed, the same samples will be in the sequence of datasets, so the test accuracies at the final time step are comparable across methods.

\textbf{Modularity} Users have the flexibility to input different imaging datasets, introduce new building blocks, construct different sequences of shifts from the building blocks, mix and match the learning approach on historical data and the fine-tuning approach at the final step, customize the layers that are fine-tuned or separate in the joint model, and run new methods on sequences constructed with our benchmark. More details on these modular components are in Appendix~\ref{app:modular}.

A limitation of this study is the constructed set of synthetic sequences is not a comprehensive collection of potential temporal distribution shifts. While we validate the conclusions from our synthetic experiments with a real-world sequence on the Portraits dataset, additional work with other real-world datasets can help inspire the creation of other synthetic sequences and bridge the gap between synthetic and real-world sequences. In addition, we focus on overall test accuracy as the metric that is evaluated at the final step. For real-world applications, other metrics, such as accuracy in particular sub-populations, may be more important. The results from our benchmark also call for more in-depth understanding of why a good initialization can be learned without leveraging the sequential nature, how parameters change over time for sequential fine-tuning and joint modeling approaches, and what would be optimal strategies for handling different types of shifts over time.

\bibliographystyle{plainnat}
\bibliography{references}

\newpage

\appendix

\section{Examples of Benchmark Sequences}

\label{app:image_examples}

This section provides examples to illustrate the sequences of shifts that can be constructed with our benchmark as discussed in Section~\ref{sec:benchmark_construction}.

\subsection{Examples of Building Blocks}

\begin{figure}[htbp]
    \centering
    \begin{minipage}[t]{.6\textwidth}
        \centering
        \includegraphics[width=\textwidth,trim={0 .2cm 0cm .2cm},clip]{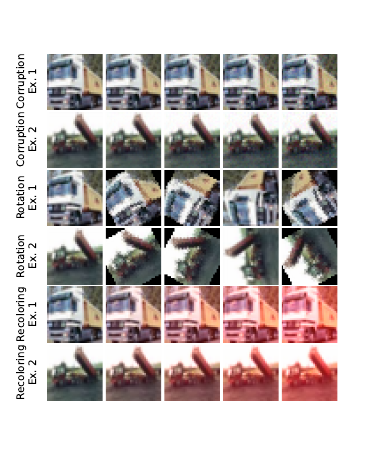}
    \end{minipage}
    \caption[Illustration of corruption, rotation, and recoloring applied repeatedly to two images.]{Illustration of corruption, rotation, and recoloring applied repeatedly to two images. Leftmost images are the original images from CIFAR-10. Transformations are applied repeatedly going to the right.}
    \label{fig:crt_repeat_examples}
\end{figure}

To illustrate the building blocks introduced in Section~\ref{sec:shift_building_blocks}, we show some examples using CIFAR-10 and CIFAR-100: 

\textbf{Input-level shift} Figure~\ref{fig:crt_repeat_examples} illustrates the 3 input-level shifts: corruption, rotation, and recoloring. These transformations are applied repeatedly to the same two images. The effects of corruptions are not visibly apparent because we chose to introduce only a small amount of noise. When large corruptions are applied, the images become impossible to classify, and accuracy is low even when a model is learned from a large number of samples. Such a set-up is not ideal for studying distribution shift because no method can address a shift that is not learnable. Rotations that are not multiples of 90 degrees introduce black corners and cut off the original corners. The recoloring transformation makes images appear more red.

\textbf{Output-level shift} After the first label flip in a sequence, the following label pairs are swapped: plane and truck, car and ship, bird and horse, cat and frog, and deer and dog. After the second label flip in a sequence, images that were labelled truck in the previous step would now be labelled horse. That means planes would now be labelled as horses.

\begin{figure}[htbp]
    \centering
    \begin{minipage}[t]{.6\textwidth}
        \centering
        \includegraphics[width=\textwidth,trim={0 .2cm 0cm .2cm},clip]{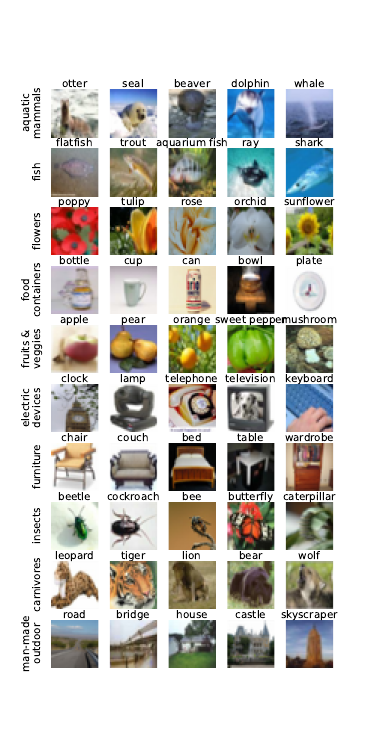}
    \end{minipage}
    \caption[Order of sub-populations in first 10 label classes in CIFAR-100.]{Order of sub-populations in first 10 label classes in CIFAR-100. The first two sub-populations are present at step 0. The other three sub-populations are introduced in the order shown at steps 1, 2, and 3.}
    \label{fig:cifar_subpop_order_1}
\end{figure}

\begin{figure}[htbp]
    \centering
    \begin{minipage}[t]{.6\textwidth}
        \centering
        \includegraphics[width=\textwidth,trim={0 .2cm 0cm .2cm},clip]{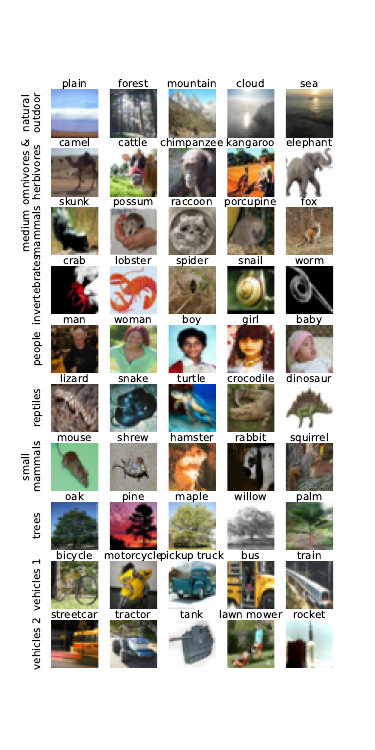}
    \end{minipage}
    \caption[Order of sub-populations in second 10 label classes in CIFAR-100.]{Order of sub-populations in second 10 label classes in CIFAR-100. The first two sub-populations are present at step 0. The other three sub-populations are introduced in the order shown at steps 1, 2, and 3.}
    \label{fig:cifar_subpop_order_2}
\end{figure}

\textbf{Intermediate-level shift} For conditional rotations, the rotated images look similar to those in Figure~\ref{fig:crt_repeat_examples}, except the rotations are opposite directions for some label classes. The black corners in different directions make it easier to distinguish the label classes when the rotations are not multiples of 90 degrees. For sub-population shifts, Figures~\ref{fig:cifar_subpop_order_1} and~\ref{fig:cifar_subpop_order_2} show the order of sub-classes introduced at each step in a sequence of sub-population shifts on CIFAR-100. For example, the vehicle label starts with bicycles and motorcycles. The 3 steps then add pick-up trucks, buses, and trains. The mammal label starts with mouses and shrews. Then, the 3 steps add hamsters, rabbits, and squirrels.

\newpage
\subsection{Examples from Sequences}

\begin{figure}[htbp]
    \centering
    \begin{minipage}[t]{.6\textwidth}
        \centering
        \includegraphics[width=\textwidth,trim={0 .2cm 0cm .2cm},clip]{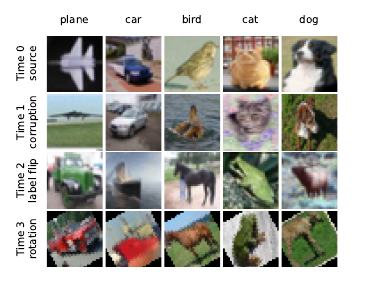}
    \end{minipage}
    \caption[Images in CIFAR-10 shift sequence: corruption, label flip, and rotation.]{Images in CIFAR-10 shift sequence: corruption, label flip, and rotation. At each step, a new shift is added on top of the shifts in the previous rows.}
    \label{fig:cifar_clr_visual}
\end{figure}

\begin{figure}[htbp]
    \centering
    \begin{minipage}[t]{.6\textwidth}
        \centering
        \includegraphics[width=\textwidth,trim={0 .2cm 0cm .2cm},clip]{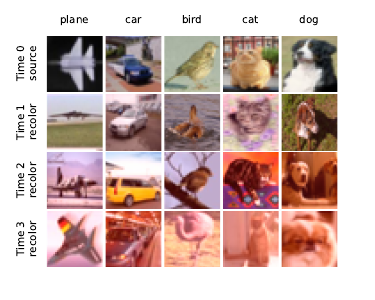}
    \end{minipage}
    \caption{Images in CIFAR-10 shift sequence with progressively more red tinting.}
    \label{fig:cifar_TTT_visual}
\end{figure}

\textbf{Fixed-length sequences of different types of shift} In Section~\ref{sec:sft_synthetic_setup}, we construct several sequences of synthetic shifts on CIFAR-10 and CIFAR-100 images. To illustrate, Figure~\ref{fig:cifar_clr_visual} shows a random sampling of images from the CIFAR-10 shift sequence with corruptions, label flips, and rotations. Figure~\ref{fig:cifar_TTT_visual} shows a random sampling of images from the shift sequence with additional red tinting applied at each step. At each step, the images are equally divided between the label classes and randomly drawn from the training samples. The images at each time step are not transformed versions of images at the previous time step. However, because they are random draws from the same CIFAR training dataset, there is some overlap between the images at different time steps. Since there are 50,000 samples in the training split of CIFAR-10 and approximately 5,000 training samples at each step, each sample is drawn with probability .1 at each step. The draws are independent at different time steps, so for a pair of time steps, a sample is drawn at both steps with .01 probability. In expectation then, any two time steps have 500 images in common before transformations are applied. This overlap is reasonable for real-world applications. As an example, in medical data, if predictions are made at each visit to a doctor, patients who routinely visit the doctor will be include in the datasets at multiple time points. The 5,000 test samples at the final time step are drawn from the 10,000 samples in the CIFAR test dataset, so there is no overlap between the final test set and the training sets at any time point. We also follow the data augmentation practices recommended in \citet{krizhevsky2017imagenet} and randomly modify each image with horizontal flips, center crops, and rotations up to 10 degrees to prevent over-fitting.

\newpage
\textbf{Variable-length sequences of rotations} We also construct many variable-length sequences of 30-degree rotations on CIFAR-10. We hold the total amount of data constant across all steps at 20k so longer sequences do not benefit from seeing more data. Thus, we keep 4k training samples at the final step and distribute the other 16k samples across the other steps as shown in Table~\ref{tab:rotation_seq_sample_sizes}. This means longer sequences may be more challenging to learn because there is less data similar to the final distribution. If data is collected at a steady rate and more data becomes available over time, it may also be interesting to design variable-length sequences with the same amount of data at each step. In our case, we find that having increasingly more data does not allow for a fair comparison of performance across different sequence lengths, so we keep the total sample size constant.

\begin{table}[htbp]
    \centering
    \begin{tabular}{cl}
    \toprule
    \# time steps & Sample size sequence (Thousands) \\
    \midrule
    2 & 16, 4 \\
    3 & 10, 6, 4 \\
    4 & 6, 4, 6, 4 \\
    5 & 4, 4, 4, 4, 4 \\
    6 & 4, 3, 3, 3, 3, 4 \\
    7 & 4, 2, 3, 2, 3, 2, 4 \\
    8 & 4, 2, 2, 2, 2, 2, 2, 4 \\
    \bottomrule
    \end{tabular}
    \caption[Number of samples at each time step in rotation sequence in thousands.]{Number of samples at each time step in rotation sequence in thousands. All sequences have 4k samples at the final time step and 16k samples total across the other time steps.}
    \label{tab:rotation_seq_sample_sizes}
\end{table}

\begin{figure}[h]
    \centering
    \begin{minipage}[t]{.7\textwidth}
        \centering
        \includegraphics[width=\textwidth,trim={0 .2cm 0cm .2cm},clip]{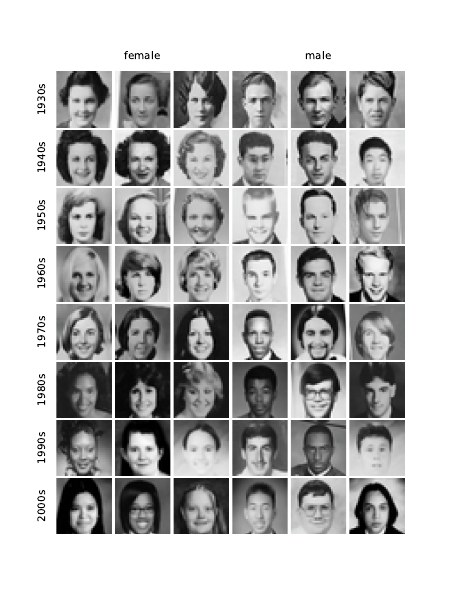}
    \end{minipage}
    \caption[Images in Portraits sequence from 1930's to 2000's.]{Images in Portraits sequence from 1930's to 2000's. Left 3 columns are female. Right 3 columns are male.}
    \label{fig:portraits_visual}
\end{figure}

\textbf{Portraits sequence} Table~\ref{tab:real_sample_sizes} lists the sample sizes at each step of the Portraits sequence. The samples are split 80/20 between training and validation for all time periods, except the last period, which is split 40/10/50 among training, validation, and test. Figure~\ref{fig:portraits_visual} shows a random sampling of portraits across the decades in the real-world dataset in Section~\ref{sec:real_world_seq}. These images reflect changes in demographics and hairstyles over the decades, as well changes in image backgrounds and lighting.

\begin{table}[htbp]
    \centering
    \begin{tabular}{cc}
    \toprule
    Time period & Sample size \\
    \midrule
    1930s & 2686 \\
    1940s & 4478 \\
    1950s & 3810 \\
    1960s & 4776 \\
    1970s & 5176 \\
    1980s & 5442 \\
    1990s & 5307 \\
    2000s (2000 - 2013) & 5514 \\
    \bottomrule
    \end{tabular}
    \caption[Sample sizes in Portraits sequence.]{Sample sizes in the Portraits sequence.}
    \label{tab:real_sample_sizes}
\end{table}

\clearpage
\section{Quantifying Shift}

\label{app:quantifying_shift}

To get a quantitative perception of how much shift is introduced over the sequence, we measure the distance between each time step and the final time step.

\subsection{Definition of Metric Quantifying Shift}
We use the Wasserstein-2 metric to quantify the distances, but first we perform PCA on the 784 pixels in the 32 x 32 training images from all time steps and select the top components that explain 95\% of the variance or at most 100 components. When the dimensions are reduced, matching samples from different steps to compute the Wasserstein distances is more feasible. By summarizing the raw pixels, this metric captures how the model perceives the input images.

\textbf{Metric for covariate shift} To quantify covariate shift between $\mathbb{P}_t\left(X\right)$ and $\mathbb{P}_T\left(X\right)$, we compute the Wasserstein-2 distance between the PCA representation of the training images from time $t$ and the PCA representation of the test images from time $T$. The empirical Wasserstein-2 distance between two sets of samples $\mathbf{X}_0 = \left\{\mathbf{x}_0^{\left(i\right)}\right\}_{i=1}^n$ and $\mathbf{X}_1 = \left\{\mathbf{x}_1^{\left(j\right)}\right\}_{j=1}^m$ is defined as 
\begin{align}
    W_2\left(\mathbf{X}_0, \mathbf{X}_1\right) &= \left(\min_{\mathbf{T} \in \mathbb{R}^{n \times m}_+} \sum_{i=1}^n \sum_{j=1}^m T_{i,j} \lVert \mathbf{x}_0^{\left(i\right)} - \mathbf{x}_1^{\left(j\right)} \rVert_2 \right)^{\frac{1}{2}} \\
    & \text{s.t. } \mathbf{T} \mathbf{1}^{m \times 1} = \mathbf{1}^{n \times 1}, \mathbf{1}^{1 \times n} \mathbf{T} = \mathbf{1}^{1 \times m}
\end{align}
$\mathbf{T}$ is the optimal transport matrix from $\mathbf{X}_0$ to $\mathbf{X}_1$. We compute the Wasserstein-2 distance using the Python optimal transport library \citep{flamary2021pot}.

To make this quantity more interpretable, we divide the computed distances by the Wasserstein-2 distance between the training and test images at the final time step. The metric we use for covariate shift is
\begin{equation}
    W_X^{t, T} = \frac{W_2\left(\mathbf{X}_t^{tr}, \mathbf{X}_T^{te}\right)}{W_2\left(\mathbf{X}_T^{tr}, \mathbf{X}_T^{te}\right)} \label{eq:cov_shift_metric}
\end{equation}
where $\mathbb{P}_t^{tr}$ and $\mathbf{X}_t^{tr}$ are the training distribution and training samples at time $t$, and $\mathbb{P}_T^{te}$ and $\mathbf{X}_T^{te}$ are the test distribution and test samples at time $T$. The denominator is the empirical distance when the test set is in-distribution. When this normalized metric is above 1, it indicates more shift than what is expected when the test set is in-distribution.

\textbf{Metric for conditional shift} To quantify conditional shift between $\mathbb{P}_t\left(X \vert Y\right)$ and $\mathbb{P}_T\left(X \vert Y\right)$, we separate the images by label class, compute the Wasserstein-2 distance separately for each class, and take the average weighted by the frequency of the label class $\mathbb{P}_T\left(Y\right)$. That is, we define the average empirical Wasserstein-2 distance between the conditional distributions of two sets of samples $\left(\mathbf{X}_0, \mathbf{Y}_0\right) = \left\{\left(\mathbf{x}_0^{\left(i\right)}, y_0^{\left(i\right)}\right)\right\}_{i=1}^n$ and $\left(\mathbf{X}_1, \mathbf{Y}_1\right) = \left\{\left(\mathbf{x}_1^{\left(j\right)}, y_1^{\left(j\right)}\right)\right\}_{j=1}^m$ as
\begin{align}
    W_2\left(\left(\mathbf{X}_0, \mathbf{Y}_0\right), \left(\mathbf{X}_1, \mathbf{Y}_1\right)\right) = \sum_{l = 1}^L \frac{\sum_{j=1}^m \mathds{1}\left\{y_1^{\left(j\right)} = l\right\}}{m} \left(\sum_{i: y_0^{\left(i\right)} = l} \sum_{j: y_1^{\left(j\right)} = l} T^{l*}_{i,j} \lVert \mathbf{x}_0^{\left(i\right)} - \mathbf{x}_1^{\left(j\right)} \rVert_2 \right)^{\frac{1}{2}}
\end{align}
where $l$ indexes into the $L$ label classes and $\mathbf{T}^{l*}$ is the transport matrix optimized separately for each label class. Similarly, the metric we use for conditional shift is also normalized by the distance between samples from the same final distribution so a metric above 1 indicates more shift is present than what would be seen for in-distribution evaluation:
\begin{equation}
    W_{X \vert Y}^{t, T} = \frac{W_2\left(\left(\mathbf{X}_t^{tr}, \mathbf{Y}_t^{tr}\right), \left(\mathbf{X}_T^{te}, \mathbf{Y}_T^{te}\right)\right)}{W_2\left(\left(\mathbf{X}_T^{tr}, \mathbf{Y}_T^{tr}\right), \left(\mathbf{X}_T^{te}, \mathbf{Y}_T^{te}\right)\right)} \label{eq:cond_shift_metric}
\end{equation}

We do not measure changes in $\mathbb{P}\left(Y\right)$ since they tend to have less effect on predictive performance and we distributed the samples equally across all label classes when constructing the synthetic sequences. For the Portraits sequence, changes in $\mathbb{P}\left(Y\right)$ can be quantified easily by measuring the change in the proportion of male or female portraits. 

The metrics above are similar to the Frechet inception distance, which measures the Wasserstein-2 distance between real and generated images from GANs using the representation in the final layer of an Inception v3 model \citep{heusel2017gans}. A distance metric like Wasserstein-2 makes more sense for measuring shift than KL divergence because the KL divergence from $\mathbb{P}$ to $\mathbb{Q}$ is undefined when some images in $\mathbb{P}$ never occur in $\mathbb{Q}$, as is often the case for distribution shift. We use the PCA representation to remain agnostic of the model architecture.

\subsection{Measuring Shift in Benchmark Sequences}

\begin{table}[htbp]
    \centering
    \begin{tabular}{lccc}
    \toprule
    Sequence & $W_X^{0, 3}$ & $W_X^{1, 3}$ & $W_X^{2, 3}$ \\
    \midrule
    Corruption, label, rotation & 1.69 & 1.69 & 1.69 \\
    Rotation, corruption, label & 1.65 & 1.00 & 1.00 \\
    Conditional rotation x3 & 1.12 & 1.55 & 1.53 \\
    Red tinting x3 & 1.54 & 1.28 & 1.08 \\
    Sub-population shift x3 & 1.01 & 1.00 & 1.00 \\
    \midrule
    Set-up & $W_{X \vert Y}^{0, 3}$ & $W_{X \vert Y}^{1, 3}$ & $W_{X \vert Y}^{2, 3}$ \\
    \midrule
    Corruption, label, rotation & 1.68 & 1.68 & 1.61 \\
    Rotation, corruption, label & 1.64 & 1.13 & 1.12 \\
    Conditional rotation x3 & 1.13 & 1.51 & 1.47 \\
    Red tinting x3 & 1.46 & 1.24 & 1.06 \\
    Sub-population shift x3 & 1.00 & 1.01 & 1.00 \\
    \bottomrule
    \end{tabular}
    \caption[Quantifying covariate and conditional shift at each step in the synthetic sequences.]{Quantifying covariate and conditional shift (top and bottom, respectively) at each step in the synthetic sequences with the metrics defined in Equations~\ref{eq:cov_shift_metric} and~\ref{eq:cond_shift_metric} based on the Wasserstein-2 distance from the final distribution. Distances above 1 indicate more shift than in-distribution evaluation.}
    \label{tab:quantify_shifts_synthetic}
\end{table}

Table~\ref{tab:quantify_shifts_synthetic} shows these normalized distances for the synthetic sequences. We can make the following observations:

\textbf{Time steps are progressively more similar to final step for rotation, corruption, then label flip and recoloring sequences} For the rotation, corruption, and label shift sequence (RCL) and the recoloring sequence (TTT), the distances reflect how the images are getting more similar to the final distribution as the sequence progresses to later steps. These sequences obey Equation~\ref{eq:shift_increases_dist}---the specification that distances tend to be larger for time steps that are farther apart in our definition for sequential distribution shift---with a negligible $\epsilon$. In the sequence with rotations, corruptions, then label flips, there is no difference between $W_X^{1,3}$ and $W_X^{2,3}$ as well as between $W_{X \vert Y}^{1,3}$ and $W_{X \vert Y}^{2,3}$. This is because the corruption introduced at step 2 does not introduce much change. This matches what we saw visually in Figure~\ref{fig:crt_repeat_examples}. Some of the effects of the label flip are captured by the metric for conditional shift.

\textbf{Source distribution is more similar to final distribution with 90 degrees of rotation than intermediate distributions} For the conditional rotation sequence, the intermediate distributions have black corners, similar to the last row in Figure~\ref{fig:cifar_clr_visual}, whereas after 90 degrees of rotations, images at the final time point are able to fill up the entire field of view. The absence of black corners in the source and final distribution make them more similar.

\textbf{Sequence of sub-population shifts shows little covariate and conditional shift} Even though the sub-classes for each label class are very different from each other, the metrics for both covariate shift and conditional shift at all 3 steps are around 1. This suggests that images from different sub-classes may have similar color frequencies. For instance, they may be on similar backgrounds or share some defining properties that make the sub-classes appear fairly similar to each other. 

\textbf{Time steps are generally progressively more similar to final step for real-world Portraits sequence} The metrics in Table~\ref{tab:quantify_shifts_real} suggest the 1940s to 1960s may have been characterized by larger differences from the 2000s than the 1970s to 1990s. From the 1970s to the 1990s, the distance to the 2000s gradually decreases. The 1930s are an exception where the metrics suggest they may be slightly more similar to the 2000s. These metrics suggest distribution shift is occurring over the decades and the Portraits dataset is a good real-world dataset for validating the conclusions from our synthetic sequences.

\begin{table}[htbp]
    \centering
    \begin{tabular}{cccc}
    \toprule
    Time period & $W_X^{19x0s, 2000s}$ & $W_{X \vert Y}^{19x0s, 2000s}$ \\
    \midrule
    1930s & 1.40 & 1.41 \\
    1940s & 1.61 & 1.68 \\
    1950s & 1.77 & 1.77 \\
    1960s & 1.62 & 1.61 \\
    1970s & 1.31 & 1.32 \\
    1980s & 1.26 & 1.27 \\
    1990s & 1.13 & 1.13 \\
    \bottomrule
    \end{tabular}
    \caption[Quantifying covariate and conditional shift at each decade in the Portraits sequence.]{Quantifying covariate and conditional shift at each decade in the Portraits sequence with the metrics defined in Equations~\ref{eq:cov_shift_metric} and~\ref{eq:cond_shift_metric} based on the Wasserstein-2 distance from the final distribution in the 2000s. Distances above 1 indicate more shift than in-distribution evaluation.
    }
    \label{tab:quantify_shifts_real}
\end{table}

In general, these metrics provide a rough estimate of the amount of shift at the input pixel level. While humans would say images at later time points are closer to the final distribution, the shifts may not appear this way to a computer vision model on the pixel level. However, we note that measuring the Wasserstein-2 distances between the covariate and conditional distributions at different time steps may not fully capture which shifts are more challenging to learn. Classifying the images correctly is harder when images from one label class become more similar to images from another label class at the previous time step. A metric that captures this would reflect a much larger distance for label flips, for example. However, we also need to be cautious about measuring what needs to changed in the model to address the shift. Label flips are relative relatively easy to address as only the output layer needs to be retrained. In contrast, intermediate-level shifts are more difficult to address, so we would expect the conditional rotation and sub-population shifts to have larger metrics. A model-based representation that measures the distance between outputs from intermediate model layers for different label classes at different time steps may be another way to quantify shifts since the distance between representations may be informative about whether the model would confuse different label classes over time.

\clearpage
\section{Image Classification Models}
\label{app:image_class_models}

This appendix provides more details on the 12 model architectures used in our experiments. We ran experiments with 3 common image classification architectures to vary the model capacity. From smallest to largest, these are ConvNets, DenseNets, and ResNets. For each architecture, we also varied the model depth. Although there are standard versions of some of these models that are pre-trained on ImageNet, we are interested in studying the effects of training on limited data, so our models are initialized from scratch.

\textbf{ConvNet} A ConvNet is a standard convolutional neural network. It starts with an input convolution layer that maps to 64 output channels. Then, convolutional blocks may be added in the middle. A block in a ConvNet contains only a single convolution layer followed by batch normalization. We use Conv-$n$ to denote a ConvNet with $n$ blocks. The intermediate layer sizes are 128 channels for Conv-2, 128 and 256 channels for Conv-3, and 64, 128, and 256 channels for Conv-4. The final convolutional layer outputs 512 channels. Finally, average pooling is applied to the output from the last convolutional block, and an output fully connected layer is applied to predict the probability of each label class.

\textbf{DenseNet} A DenseNet is a type of convolutional neural network where the output channels from each layer are also passed forward to each subsequent layer within the same dense block \citep{huang2017densely}. The output from each layer is concatenated to its inputs, forming increasingly larger input dimensions in later layers. The dense blocks follow this structure. There are also an input convolution layer and an output fully connected layer. The layers in the dense blocks grow in output dimensions by 32 each time, starting at 64. Dense-1 has 16 layers in its dense block, Dense-2 has 6 and 16 layers, Dense-3 has 6, 12, and 16 layers, and Dense-4 has 6, 12, 24, and 16 layers. Dense-4 in our experiments is the standard DenseNet-121 architecture.

\textbf{ResNet} A ResNet is also a type of convolutional neural network where the output is added to its input, i.e. a layer outputs $f\left(x\right) + x$, where $f$ is the standard convolutional layer \citep{he2016deep}. We create residual blocks with the same number of channels as the blocks in ConvNets. Each residual block contains 2 basic blocks, which each contain 2 convolution layers. The first convolution layer in each residual block maps from the number of input channels to the number of output channels for that residual block. The remaining 3 convolution layers map from the number of output channels to the number of output channels. When input and output dimensions differ, the residual block also has a convolution layer with kernel size 1 to downsample the input image to match the output size so they can be added in the forward pass. The ResNet-4 model in our experiments is the official ResNet-18 architecture. All of these architecture designs are shown in Table~\ref{tab:model_architectures}.

\begin{table}[htbp]
    \centering
    \begin{tabular}{ccc}
    \toprule
    Model & Size sequence & \# parameters \\
    \midrule
    Conv-1 & 512 & .3M \\
    Conv-2 & 128, 512 & .7M \\
    Conv-3 & 128, 256, 512 & 1.6M \\
    Conv-4 & 64, 128, 256, 512 & 1.6M \\
    \midrule
    Dense-1 & 16 & 1.2M \\
    Dense-2 & 6, 16 & 1.7M \\
    Dense-3 & 6, 12, 16 & 3.0M \\
    Dense-4 & 6, 12, 24, 16 & 6.9M \\
    \midrule
    Res-1 & 512 & 7.4M \\
    Res-2 & 128, 512 & 8.2M \\
    Res-3 & 128, 256, 512 & 10.9M \\
    Res-4 & 64, 128, 256, 512 & 11.0M \\
    \bottomrule
    \end{tabular}
    \caption[Size of image classification models.]{Size of image classification models. For ConvNets and ResNets, the size sequence is the number of output channels from each layer. The first block always has 64 input channels. For DenseNets, the size sequence is the number of dense layers in each block. In each dense block, the number of output dimensions starts at 64 and grows by 32 each layer. Number of parameters is for 10 label classes.}
    \label{tab:model_architectures}
\end{table}

\clearpage
\section{Additional Details on Benchmark}

\label{app:benchmark_details}

In this appendix, we provide additional details on how some of the methods are implemented and how the modular components in the benchmark can be combined to run a versatile suite of experiments.

\subsection{Additional Details on Benchmark Methods}
\label{app:benchmark_methods}

\textbf{IRM} We follow the minimal implementation suggested by \citet{arjovsky2019invariant}. The IRM objective is
\begin{equation}
    \min \sum_{t=0}^T \mathcal{L}_t\left(\theta\right) + \lambda \lVert \nabla_{w \vert w = 1.0} \mathcal{L}_t\left(w \cdot \theta\right) \rVert^2
\end{equation}
The original objective treats $\theta$ as the invariant feature representation and $w$ as the invariant classifier. However, they claim that setting $w = 1.0$ and treating $\theta$ as the entire classifier is sufficient. The minimal implementation then computes the penalty term by taking the gradient $g_1$ of the loss for a sample with respect to the dummy $w$, taking the gradient $g_2$ of the loss for another sample with respect to $w$, and then summing the product $g_1 \cdot g_2$ across pairs of samples. The Wild-Time benchmark also performs IRM on multiple source domains. In their set-up, the time intervals they create to define each domain overlap, so the domains contain overlapping settings. We keep the time periods distinct so the domains do not share samples.

\textbf{DRO} The objective minimizes the worst loss across all time steps: 
\begin{equation}
    \min_{\theta} \max_{t \in \left\{0, \ldots, T\right\}} \mathcal{L}^{\left(\theta\right)}_t
\end{equation}
At each gradient descent step, the loss is computed for a separate batch from each time step. Back-propagation is performed only on the batch with the largest loss.

\textbf{Sequential fine-tuning} When training a single dataset, many optimizers decrease the learning rate over training epochs so the model can get closer to convergence. When training over a sequence of datasets, it may also be beneficial to decay the learning rate at each time step to prevent the model from straying too far from increasingly better initial parameters. As such, we try 3 different learning rate schedules across the time steps: 1) No decay: The best learning rate is selected independently at each time step. 2) Linear decay: A constant factor is subtracted from the learning rate at each time step. The sequence of learning rates is determined by the initial rate and the constant factor. This pair is selected by hyper-parameter tuning. 3) Exponential decay: The learning rate is multiplied by a constant factor between 0 and 1 at each time step. The learning rate sequence is defined and selected in the same way as linear decay.

\textbf{Elastic weight consolidation} The objective is
\begin{equation}
    \mathcal{L}_T^{\theta} + \sum_{t=0}^{T-1} \frac{\lambda}{2} F_t \left(\theta_T - \theta_t\right)^2
\end{equation}
where $F$ is the Fisher information matrix for the parameters at the previous time step \citep{kirkpatrick2017overcoming}. $F$ captures the importance of each parameter. The regularization has a similar effect to learning rate decay since important parameters are not updated as much. Other continual learning approaches include synaptic intelligence and averaged gradient episodic memory \citep{zenke2017continual,chaudhry2018efficient}. A specific approach to continual learning is incremental learning, which is specifically designed to adapt to the introduction of new label classes \citep{castro2018end,wu2019large}.

\begin{figure}[htbp]
    \centering
    \begin{minipage}[t]{.37\textwidth}
        \centering
        \includegraphics[width=\textwidth]{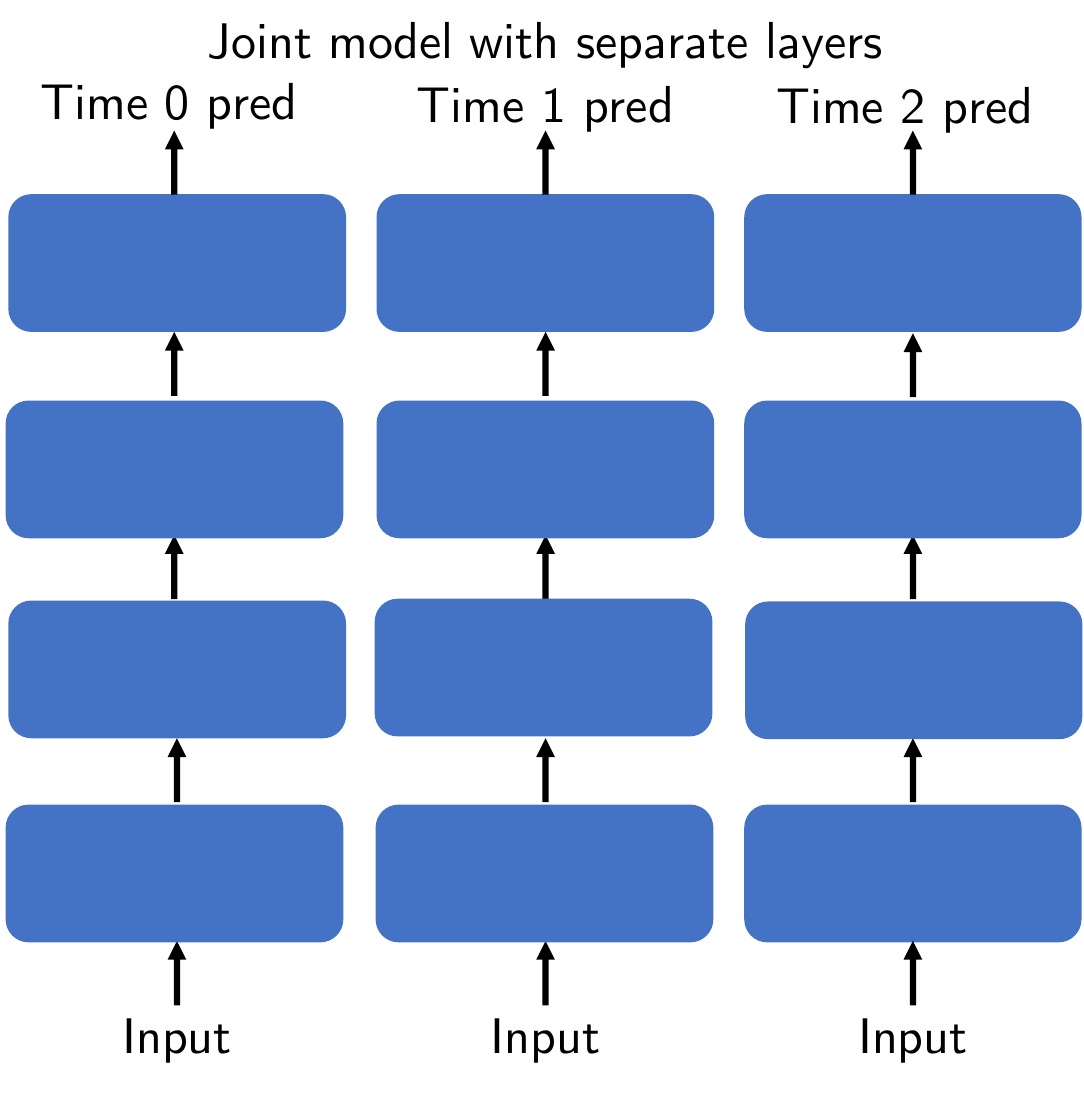}
    \end{minipage}
    \centering
    \begin{minipage}[t]{.37\textwidth}
        \centering
        \includegraphics[width=\textwidth]{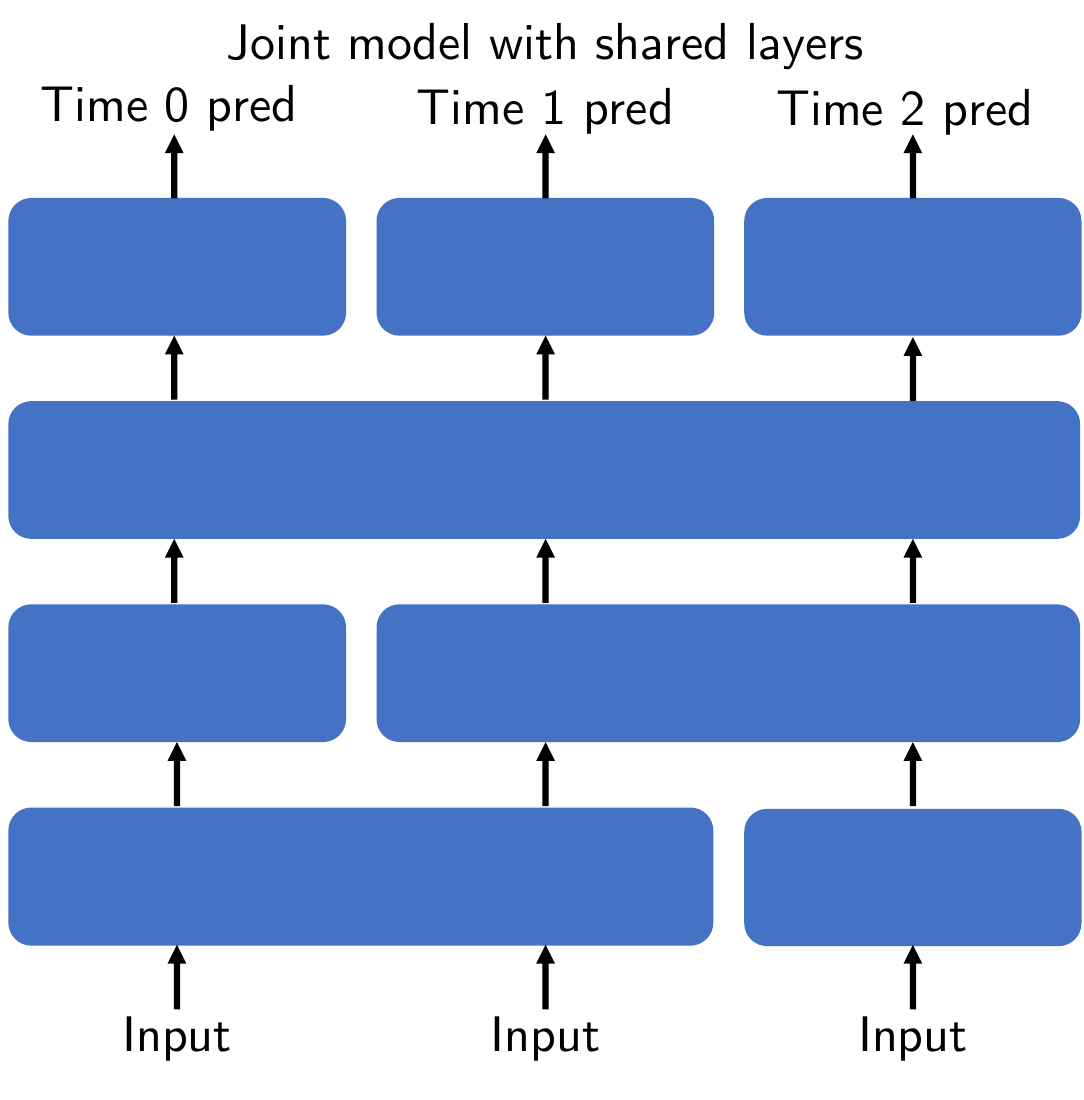}
    \end{minipage}
    \centering
    \begin{minipage}[t]{.24\textwidth}
        \centering
        \includegraphics[width=\textwidth]{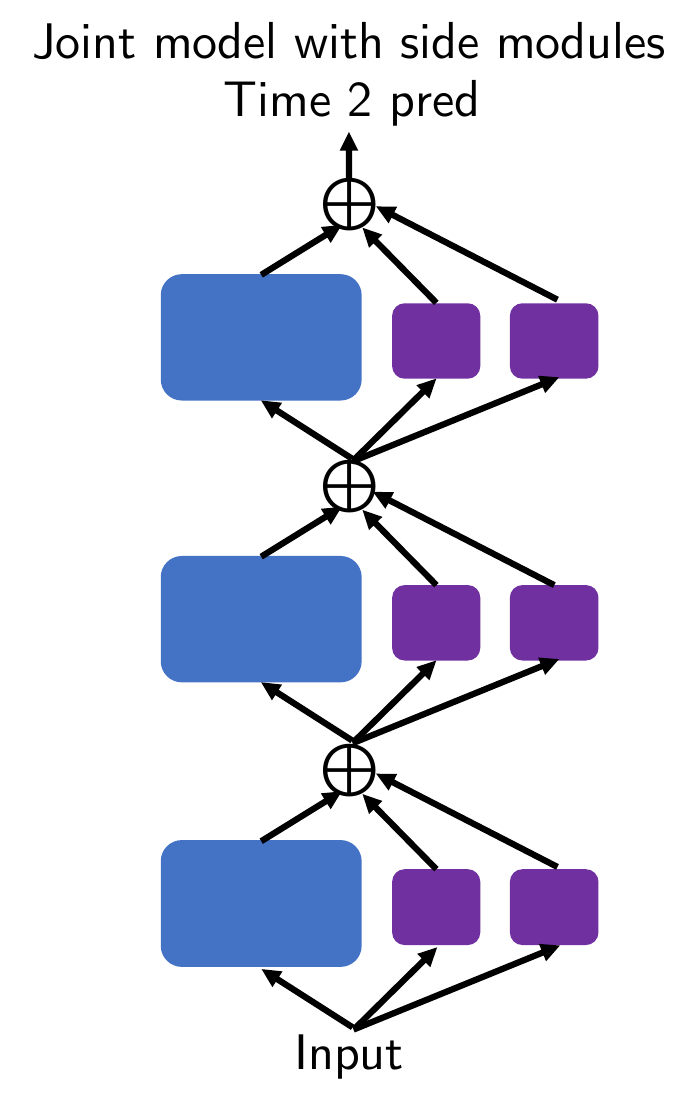}
    \end{minipage}
    \caption[Variants of the joint model.]{Variants of the joint model. Left: All layers are separate for each time step. Middle: Some layers are shared across time steps. Right: Base modules at $t = 0$ and side modules at later time steps.}
    \label{fig:joint_model_diagrams}
\end{figure}

\textbf{Joint model} The standard joint model that has separate modules at each time step is shown on the left of Figure~\ref{fig:joint_model_diagrams}. The modules are related by applying L2 regularization on the difference between parameters at adjacent time steps. We also considered tuning the amount of L2 regularization on the weights, the dropout, the learning rate, and the weight on the loss at the final time step relative to the weight on the losses at previous time steps. After some preliminary exploration, we found that tuning these other hyper-parameters did not make a large difference, so we kept them at fixed values and focused on tuning the L2 regularization on the difference between adjacent time steps. We found that giving the loss at the final time step 3 times the weight of the loss from a single previous time step tends to lead to better performance at the final time step than giving the final time step equal weight.

Our implementation of the model can also be use to share layers across some time points as shown in the middle panel of Figure~\ref{fig:joint_model_diagrams}. The layers that are shared can be selected based on the types of shifts introduced at each step. For more complicated shifts, potential heuristics for determining which layers to share could be developed. For instance, metrics similar to those from surgical fine-tuning could be useful, or layers at adjacent time steps could be merged by testing whether their outputs are similar. The structure of a joint model with side modules is shown in the right panel of Figure~\ref{fig:joint_model_diagrams}. At each layer, the outputs from the original block and all side modules leading up to the prediction time point are summed. The prediction for time 0 would only use the original blocks, and at time 1, only the original block and the first side module in each layer are summed.

\subsection{Additional Details on Modular Components}
\label{app:modular}

Our benchmark provides modular components for constructing sequences of dataset, combining methods for learning from historical data and adapting to the final distribution, and selecting different flavors of fine-tuning or joint models.

\textbf{Modular sequence construction} Users can specify the name of the dataset to draw images from, the types of shifts at each step, a list of sample sizes for each step, and the size of the test set at the final time step. For datasets, the repository currently supports CIFAR-10, CIFAR-100, and ImageNet. Users may also add their own script for loading other datasets, and the building blocks in our benchmark can be applied to these images from other datasets. The repository gives users the flexibility to create various sequences of synthetic shifts by combining the building blocks that are provided. Users can also choose to have multiple shifts at each step. They can also introduce other types of distribution shift. At a preliminary level, users can modify the code for the building blocks to introduce different rotation angles, different degrees of corruption, or different recoloring patterns for the channels. Users can also add other types of shifts that may connect to more real-world applications. Varying the number of steps and sample size at each step allows users to identify optimal strategies when different amounts of historical and current data are available.

\textbf{Modular methods} The methods can also be built in a modular fashion. First, users can select the family of methods they are running. Within each family, there are modular options as well. For methods that group all the domains or all historical domains, users can choose whether to run ERM, IRM, or DRO or supply another method. For the fine-tuning process, users can choose whether to run fine-tuning or side-tuning. We also provide some support for low-rank adapters that is explained in the repository. All the methods described in Section~\ref{sec:methods} are included in the repository and can be applied in a modular fashion as well.

\textbf{Fine-grained control of different methods} For the methods that fine-tune at the final step or perform sequential fine-tuning, users can also control which layers are fine-tuned at each step. One option is to fine-tune some layers and freeze the other layers. Which layers are fine-tuned can be specified manually or selected based on the surgical fine-tuning metrics. Layers can also be unfrozen gradually. For example, in linear probing then fine-tuning, the layer closest to the output is updated first before the other layers are unfrozen. This option can be specified as ``fc-all''. There are also shortcuts to unfreeze the layers from last-to-first or first-to-last. For sequential fine-tuning, beam search across multiple time steps can also be used to select the layers.

There are also many potential modifications for side-tuning and joint model variants. For side-tuning, users can specify whether to use side modules for the input and output layers. Because they are a single convolution layer and a single fully connected layer, the side modules would be the same size, and it may be better to avoid restricting those layers to the additive structure of side modules. For joint models, users can specify to share layers across specific time steps as shown in Figure~\ref{fig:joint_model_diagrams}. They can also assign different weights to the losses at different time steps. For joint models and sequential fine-tuning, the code supports alternative types of regularization towards previous models, including $L_1$ or $L_2$ regularization towards 0, $L_1$ or $L_2$ regularization towards previous weights, $L_2$ regularization towards previous weights at all time steps, or $L_2$ regularization that is weighted by the Fisher information matrix computed at previous time steps.

\clearpage
\section{Benchmark Results for Individual Model Architectures}
\label{app:full_results}

In Section~\ref{sec:results}, Table~\ref{tab:mean_results} presented the mean test accuracy across the 12 model architectures for each method for each sequence of shifts. In this appendix, we present the results for the individual model architectures. The model architectures in the tables are described in Appendix~\ref{app:image_class_models}. Tables~\ref{tab:clr_results_full}-\ref{tab:sss_results_full} show the results for the 5 fixed-length synthetic sequences, and Table~\ref{tab:portraits_results_full} shows the results for the real-world Portraits sequence. We also include results for the variable length rotation sequences in Table~\ref{tab:rotation_seq_results}. For these sequences, all results are on the largest 4-block ResNet architecture. All experiments were run on 3 Nvidia GeForce GTX 1080 Ti, 1 Tesla M40 24GB, and 8 Nvidia RTX 6000 Ada Generation GPUs.

In these tables, the top halves start with the baseline (4k), then enumerate the methods that group data from all time points, and finally include all the methods that group all the historical data before fine-tuning at the final time step. This corresponds to the top and middle sections of Table~\ref{tab:mean_results}. The bottom halves contain methods that leverage the sequential nature of historical data. The final column is the oracle (20k). As in Table~\ref{tab:mean_results}, the metrics are bolded if they are within 0.02 of the best for that architecture. Metrics are italicized if they are within 0.02 or better than the oracle (20k) for that architecture. The formatting for the means differs from that used in Table~\ref{tab:mean_results}. Here, the mean is bolded (or italicized) if the average difference between the method and the best (or oracle) is less than 1 standard deviation below 0. This difference captures when a method is best-performing or oracle-like for an individual architecture. When this difference is not significantly negative, that means the method is generally among the best-performing or oracle-like set across multiple architectures. Thus, the formatting in this appendix more accurately reflects the comparisons that are made across individual architectures. In contrast, a mean may be bolded in Table~\ref{tab:mean_results} when a method performs very well for some architectures and very poorly for other architectures if the mean is on par with the best-performing or oracle mean. We chose to use the formatting in Table~\ref{tab:mean_results} so that the formatting would be more intuitive when the means are presented in a stand-alone manner. From examining these tables with results for individual architectures, we draw some additional conclusions:

\textbf{Learning rate decay is more effective than Fisher regularization towards previous weights for sequential fine-tuning} For sequential fine-tuning, decaying the learning rate over the sequence of distributions almost always results in better performance than elastic weight consolidation. One conjecture is uneven constraints on the parameters may distort model updates in unforeseen directions. Another conjecture is the Fisher information matrix captures parameter importance at previous time steps, which makes sense for preventing catastrophic forgetting in continual learning. However, our goal is to adapt to the final distribution, and the Fisher information matrix computed on historical distributions does not reflect the current importance of the parameters.

\textbf{Smaller architectures may be sufficient for adapting to a final distribution with limited samples when oracle-level performance cannot be achieved} For the sequence of corruptions, label flips, then rotations in Table~\ref{tab:clr_results_full} and the sequence of conditional rotations in Table~\ref{tab:rrr_results_full}, convolutional networks with at least 2 layers are able to achieve similar performance compared to the larger architectures when fine-tuning to a limited sample size. Dense and residual networks tend to perform better when there is a large amount of data available from the final distribution, but that gain is not apparent when sample size is limited at the final step. That suggests the existing fine-tuning approaches may not be able to utilize the larger model capacity. For the other synthetic sequences in Tables~\ref{tab:rcl_results_full},~\ref{tab:ttt_results_full}, and~\ref{tab:sss_results_full} where oracle-level performance is achieved, because the historical data is more relevant, the larger capacity of dense and residual networks is utilized during fine-tuning. For the Portraits sequence, convolutional networks are also able to achieve the same accuracy as the larger architectures.

\textbf{Larger architectures may have sufficient capacity to generalize across distributions} For the sequence of corruption, label flip, then rotation in Table~\ref{tab:clr_results_full} and the sequence of conditional rotations in Table~\ref{tab:rrr_results_full}, convolutional networks fit with any of the three methods that do not target the final distribution (ERM, IRM, and DRO) underperform compared to the baseline trained only on 4k samples from the final distribution. In contrast, dense networks fit with any of these three methods attain similar accuracies to the baseline, and residual networks achieve similar if not better performance than the baseline. We hypothesize large model architectures may have the capacity to learn multiple distributions, classify which time step a sample is from, and make predictions based on the distribution for that time step. This hypothesis could be tested by examining whether different parts of the intermediate outputs are activated at different time steps.

\textbf{Leveraging the sequential nature of historical datasets is unnecessary for sequences of rotations of any length} For all sequence lengths in Table~\ref{tab:rotation_seq_results}, some variant of learning from all historical data and fine-tuning on the final distribution performs just as well as sequential fine-tuning or learning a joint model. (For length-2 sequences, sequential fine-tuning and elastic weight consolidation are essentially fine-tuning but with more hyper-parameters that can be tuned.) In general, as the sequences became longer, performance deteriorated because the same number of historical samples was spread out over distributions that were farther away. Length-4 and length-7 sequences had slightly better performance at the final time step because the images were rotated 90 and 180 degrees, respectively. That means the full image is available and there may be slightly less shift compared to the source distribution.

\begin{table}[htbp]
    \centering
    \begin{tabular}{lcccccccccc}
    \toprule
    Model & 4k & ERM & IRM & DRO & FT & LP-FT & I-FT & D-FT & ST-1 & ST-B \\
    \midrule
    Conv-1 & .47 & .36 & .41 & .36 & .49 & .50 & .47 & \textbf{.51} & .49 & --- \\
    Conv-2 & .51 & .44 & .48 & .39 & .53 & .55 & \textbf{.58} & \textbf{.56} & .51 & --- \\
    Conv-3 & .57 & .45 & .42 & .41 & .56 & \textbf{.57} & \textbf{.58} & \textbf{.58} & .56 & --- \\
    Conv-4 & .54 & .41 & .43 & .41 & \textbf{.58} & .57 & \textbf{.58} & \textbf{.58} & .56 & --- \\
    \midrule
    Dense-1 & .53 & .51 & .50 & .47 & \textbf{.61} & .59 & .58 & \textbf{.59} & .51 & .55 \\
    Dense-2 & .55 & .55 & .51 & .50 & .59 & \textbf{.61} & .57 & .58 & .54 & .56 \\
    Dense-3 & .52 & .55 & .51 & .49 & \textbf{.59} & \textbf{.60} & .58 & .56 & .56 & .55 \\
    Dense-4 & .55 & .55 & .48 & .47 & .57 & .57 & .58 & .55 & .55 & .54 \\
    \midrule
    Res-1 & .57 & .58 & .52 & .52 & \textbf{.63} & \textbf{.62} & \textbf{.64} & \textbf{.62} & .53 & .59 \\
    Res-2 & .55 & .57 & .54 & .53 & .57 & .58 & .58 & .57 & .57 & .59 \\
    Res-3 & .54 & .55 & .54 & .51 & \textbf{.59} & \textbf{.59} & \textbf{.59} & \textbf{.59} & \textbf{.58} & .57 \\
    Res-4 & .53 & .58 & .57 & .51 & .60 & .59 & \textbf{.63} & .59 & .58 & .59 \\
    \midrule
    Mean & .54 & .51 & .49 & .46 & \textbf{.58} & \textbf{.58} & \textbf{.58} & \textbf{.57} & .55 & \textbf{.55} \\
    Std dev & .03 & .07 & .05 & .05 & .03 & .03 & .04 & .03 & .02 & .03 \\
    \midrule
    Model & SFT & EWC & SST-1 & SST-B & JM & JST-1 & JST-B & & & 20k \\
    \midrule
    Conv-1 & \textbf{.52} & \textbf{.51} & .49 & --- & .48 & \textbf{.52} & --- & & & .54 \\
    Conv-2 & \textbf{.57} & \textbf{.57} & .55 & --- & .54 & .54 & --- & & & .61 \\
    Conv-3 & \textbf{.59} & .57 & \textbf{.59} & --- & .57 & .56 & --- & & & .64 \\
    Conv-4 & \textbf{.60} & \textbf{.58} & .56 & --- & .56 & .56 & --- & & & .63 \\
    \midrule
    Dense-1 & \textbf{.61} & .58 & .50 & .55 & .58 & \textbf{.61} & \textbf{.60} & & & .68 \\
    Dense-2 & \textbf{.59} & .58 & \textbf{.59} & .59 & \textbf{.61} & \textbf{.60} & \textbf{.61} & & & .68 \\
    Dense-3 & .57 & .58 & \textbf{.59} & .56 & \textbf{.60} & \textbf{.61} & \textbf{.59} & & & .68 \\
    Dense-4 & \textbf{.59} & .57 & .58 & .58 & .56 & \textbf{.61} & .59 & & & .69 \\
    \midrule
    Res-1 & \textbf{.62} & .62 & .56 & .61 & .60 & .59 & .59 & & & .70 \\
    Res-2 & \textbf{.61} & .57 & .58 & \textbf{.60} & \textbf{.59} & .57 & \textbf{.59} & & & .68 \\
    Res-3 & \textbf{.59} & \textbf{.58} & \textbf{.58} & \textbf{.58} & \textbf{.59} & \textbf{.58} & .56 & & & .69 \\
    Res-4 & .60 & .58 & .60 & \textbf{.62} & .61 & .60 & .55 & & & .71 \\
    \midrule
    Mean & \textbf{.59} & \textbf{.57} & \textbf{.57} & \textbf{.57} & \textbf{.57} & \textbf{.60} & \textbf{.57} & & & .66 \\
    Std dev & .02 & .02 & .03 & .03 & .04 & .01 & .03 & & & .05 \\
    \bottomrule
    \end{tabular}
    \caption[Test accuracy at the final time step for each model architecture for the shift sequence with 6k source samples, 4k samples with corruption (C), 6k samples with C and label flip (L), 4k samples with C, L, and rotation (R).]{Test accuracy at the final time step for each model architecture for the shift sequence with 6k source samples, 4k samples with corruption (C), 6k samples with C and label flip (L), 4k samples with C, L, and rotation (R).
    
    Oracle is last column (20k). Italicized if oracle-like (within 0.02 or better than oracle) for that architecture. Bolded if oracle-like or within 0.02 of best for that architecture (row). Mean is italicized if the average difference between the method and the oracle is less than 1 standard deviation below 0. Mean is bolded if the average difference between the method and either the best or the oracle is less than 1 standard deviation below 0.}
    \label{tab:clr_results_full}
\end{table}

\begin{table}[htbp]
    \centering
    \begin{tabular}{lcccccccccc}
    \toprule
    Model & 4k & ERM & IRM & DRO & FT & LP-FT & I-FT & D-FT & ST-1 & ST-B \\
    \midrule
    Conv-1 & .47 & .17 & .22 & .26 & \textit{\textbf{.54}} & \textit{\textbf{.55}} & .52 & .48 & .50 & --- \\
    Conv-2 & .51 & .18 & .21 & .31 & \textbf{.58} & \textit{\textbf{.60}} & .57 & .56 & .58 & --- \\
    Conv-3 & .57 & .21 & .19 & .29 & \textbf{.60} & \textbf{.60} & .59 & .58 & .58 & --- \\
    Conv-4 & .54 & .19 & .20 & .28 & \textbf{.60} & \textit{\textbf{.62}} & .59 & .58 & \textbf{.60} & --- \\
    \midrule
    Dense-1 & .53 & .20 & .24 & .28 & .60 & \textit{\textbf{.66}} & .54 & .56 & .56 & .56 \\
    Dense-2 & .55 & .20 & .21 & .30 & .60 & \textbf{.66} & .59 & .56 & .60 & .61 \\
    Dense-3 & .52 & .19 & .26 & .32 & .60 & \textit{\textbf{.66}} & .58 & .56 & .61 & .63 \\
    Dense-4 & .55 & .20 & .23 & .27 & .61 & \textbf{.66} & .58 & .56 & .62 & .64 \\
    \midrule
    Res-1 & .57 & .16 & .28 & .35 & .65 & \textbf{.68} & .64 & .61 & .65 & .64 \\
    Res-2 & .55 & .17 & .20 & .32 & .62 & \textbf{.64} & .61 & .57 & \textbf{.65} & \textbf{.65} \\
    Res-3 & .54 & .18 & .46 & .31 & \textbf{.67} & \textit{\textbf{.67}} & .63 & .58 & \textbf{.66} & \textbf{.66} \\
    Res-4 & .53 & .21 & .23 & .32 & \textbf{.67} & \textbf{.67} & .65 & .60 & \textbf{.67} & .65 \\
    \midrule
    Mean & .54 & .19 & .24 & .30 & .61 & \textbf{.64} & .59 & .57 & \textbf{.63} & .61 \\
    Std dev & .03 & .01 & .07 & .02 & .04 & .04 & .04 & .03 & .04 & .04 \\
    \midrule
    Model & SFT & EWC & SST-1 & SST-B & JM & JST-1 & JST-B & & & 20k \\
    \midrule
    Conv-1 & \textit{\textbf{.53}} & .49 & .45 & --- & .45 & .51 & --- & & & .54 \\
    Conv-2 & \textbf{.58} & .57 & .54 & --- & .55 & .56 & --- & & & .61 \\
    Conv-3 & \textbf{.61} & \textbf{.60} & .59 & --- & .57 & .56 & --- & & & .64 \\
    Conv-4 & \textbf{.61} & .58 & .58 & --- & .58 & .57 & --- & & & .63 \\
    \midrule
    Dense-1 & .62 & .61 & .50 & .57 & .59 & .64 & .62 & & & .68 \\
    Dense-2 & .60 & .56 & .57 & .59 & .56 & .63 & .64 & & & .68 \\
    Dense-3 & .59 & .59 & .59 & .59 & .58 & .64 & .64 & & & .68 \\
    Dense-4 & .63 & .56 & .56 & .61 & .59 & .62 & .64 & & & .69 \\
    \midrule
    Res-1 & .65 & .62 & .58 & .58 & .63 & .65 & .63 & & & .70 \\
    Res-2 & \textbf{.64} & .63 & .63 & .62 & .62 & .62 & \textbf{.64} & & & .68 \\
    Res-3 & .64 & .62 & \textbf{.65} & .64 & .62 & .63 & .60 & & & .69 \\
    Res-4 & \textbf{.66} & .61 & \textbf{.66} & .64 & .63 & .64 & .55 & & & .71 \\
    \midrule
    Mean & .61 & .59 & .59 & .59 & .58 & .63 & .60 & & & .66 \\
    Std dev & .03 & .04 & .05 & .05 & .05 & .01 & .04 & & & .05 \\
    \bottomrule
    \end{tabular}
    \caption[Full benchmark of test accuracy at final time step for shift sequence with 6k source samples, 4k samples with rotation (R), 6k samples with R and corruption (C), and 4k samples with R, C, and label flips (L).]{Full benchmark of test accuracy at final time step for shift sequence with 6k source samples, 4k samples with rotation (R), 6k samples with R and corruption (C), and 4k samples with R, C, and label flips (L).

    Oracle is last column (20k). Italicized if oracle-like (within 0.02 or better than oracle) for that architecture. Bolded if oracle-like or within 0.02 of best for that architecture (row). Mean is italicized if the average difference between the method and the oracle is less than 1 standard deviation below 0. Mean is bolded if the average difference between the method and either the best or the oracle is less than 1 standard deviation below 0.}
    \label{tab:rcl_results_full}
\end{table}

\begin{table}[htbp]
    \centering
    \begin{tabular}{lcccccccccc}
    \toprule
    Model & 4k & ERM & IRM & DRO & FT & LP-FT & I-FT & D-FT & ST-1 & ST-B \\
    \midrule
    Conv-1 & .52 & .47 & .51 & .44 & .54 & .55 & \textit{\textbf{.59}} & .56 & .51 & --- \\
    Conv-2 & .57 & .53 & .56 & .51 & \textbf{.63} & \textbf{.62} & \textbf{.63} & \textbf{.62} & .61 & --- \\
    Conv-3 & .60 & .55 & .54 & .53 & \textbf{.63} & .62 & \textbf{.64} & \textbf{.64} & \textbf{.63} & --- \\
    Conv-4 & .60 & .55 & .57 & .50 & \textbf{.64} & \textbf{.64} & \textbf{.64} & \textbf{.62} & .61 & --- \\
    \midrule
    Dense-1 & .63 & .58 & .56 & .59 & \textbf{.66} & \textbf{.66} & .64 & \textbf{.67} & .55 & .64 \\
    Dense-2 & .59 & .60 & .58 & .59 & \textbf{.65} & \textbf{.65} & .63 & \textbf{.65} & .62 & \textbf{.64} \\
    Dense-3 & .61 & .59 & .58 & .57 & .61 & .62 & .61 & \textbf{.65} & \textbf{.64} & .63 \\
    Dense-4 & .60 & .57 & .59 & .58 & \textbf{.65} & \textbf{.65} & .63 & \textbf{.64} & \textbf{.65} & \textbf{.64} \\
    \midrule
    Res-1 & .60 & .62 & .61 & .59 & \textbf{.70} & \textbf{.69} & .65 & \textbf{.69} & .57 & .67 \\
    Res-2 & .58 & .59 & .60 & .58 & \textbf{.63} & .62 & \textbf{.64} & \textbf{.64} & .62 & .62 \\
    Res-3 & .58 & .60 & .59 & .58 & \textbf{.63} & \textbf{.63} & \textbf{.63} & \textbf{.63} & \textbf{.63} & \textbf{.63} \\
    Res-4 & .57 & .63 & .61 & .56 & .65 & .65 & .63 & .63 & .65 & \textbf{.67} \\
    \midrule
    Mean & .59 & .57 & .57 & .55 & \textbf{.63} & \textbf{.63} & \textbf{.63} & \textbf{.64} & \textbf{.62} & \textbf{.63} \\
    Std dev & .03 & .04 & .03 & .05 & .04 & .03 & .02 & .03 & .03 & .04 \\
    \midrule
    Model & SFT & EWC & SST-1 & SST-B & JM & JST-1 & JST-B & & & 20k \\
    \midrule
    Conv-1 & .56 & .57 & .53 & --- & .54 & .54 & --- & & & .60 \\
    Conv-2 & \textbf{.63} & .61 & .61 & --- & .58 & .60 & --- & & & .67 \\
    Conv-3 & \textbf{.65} & .62 & \textbf{.63} & --- & .61 & .60 & --- & & & .68 \\
    Conv-4 & \textbf{.64} & .62 & \textbf{.63} & --- & .61 & .61 & --- & & & .68 \\
    \midrule
    Dense-1 & .62 & .65 & .61 & .64 & \textbf{.66} & .60 & .60 & & & .74 \\
    Dense-2 & \textbf{.65} & \textbf{.65} & .62 & \textbf{.65} & \textbf{.66} & .62 & .59 & & & .76 \\
    Dense-3 & \textbf{.65} & .63 & \textbf{.66} & .62 & \textbf{.65} & .63 & .61 & & & .74 \\
    Dense-4 & \textbf{.65} & \textbf{.64} & .63 & \textbf{.64} & .60 & .62 & .60 & & & .74 \\
    \midrule
    Res-1 & \textbf{.68} & .66 & .52 & .59 & .66 & .62 & .63 & & & .75 \\
    Res-2 & .61 & .61 & .62 & \textbf{.64} & .62 & .61 & \textbf{.63} & & & .74 \\
    Res-3 & .61 & .60 & .61 & \textbf{.64} & \textbf{.62} & .62 & .59 & & & .74 \\
    Res-4 & .62 & .62 & .63 & .64 & .62 & .62 & .57 & & & .75 \\
    \midrule
    Mean & \textbf{.63} & .62 & \textbf{.61} & \textbf{.62} & .62 & .62 & .60 & & & .72 \\
    Std dev & .03 & .02 & .04 & .03 & .04 & .01 & .02 & & & .05 \\
    \bottomrule
    \end{tabular}
    \caption[Test accuracy at the final time step for each model architecture for the shift sequence with 6k source samples, 4k samples with 30-degree conditional rotations (r), 6k samples with 60-degree r, and 4k samples with 90-degree r.]{Test accuracy at the final time step for each model architecture for the shift sequence with 6k source samples, 4k samples with 30-degree conditional rotations (r), 6k samples with 60-degree r, and 4k samples with 90-degree r.
    
    Oracle is last column (20k). Italicized if oracle-like (within 0.02 or better than oracle) for that architecture. Bolded if oracle-like or within 0.02 of best for that architecture (row). Mean is italicized if the average difference between the method and the oracle is less than 1 standard deviation below 0. Mean is bolded if the average difference between the method and either the best or the oracle is less than 1 standard deviation below 0.}
    \label{tab:rrr_results_full}
\end{table}

\begin{table}[htbp]
    \centering
    \begin{tabular}{lcccccccccc}
    \toprule
    Model & 4k & ERM & IRM & DRO & FT & LP-FT & I-FT & D-FT & ST-1 & ST-B \\
    \midrule
    Conv-1 & .48 & \textit{\textbf{.55}} & .50 & .50 & \textit{\textbf{.58}} & \textit{\textbf{.58}} & \textit{\textbf{.58}} & \textit{\textbf{.58}} & .54 & --- \\
    Conv-2 & .55 & .60 & .59 & .54 & \textit{\textbf{.64}} & \textit{\textbf{.63}} & \textit{\textbf{.64}} & .58 & \textit{\textbf{.62}} & --- \\
    Conv-3 & .56 & \textit{\textbf{.64}} & .62 & .53 & \textit{\textbf{.65}} & \textit{\textbf{.64}} & \textit{\textbf{.67}} & .58 & \textit{\textbf{.63}} & --- \\
    Conv-4 & .58 & .53 & \textit{\textbf{.62}} & .54 & \textit{\textbf{.65}} & \textit{\textbf{.65}} & \textit{\textbf{.65}} & .57 & .61 & --- \\
    \midrule
    Dense-1 & .56 & \textit{\textbf{.69}} & .63 & .64 & \textit{\textbf{.70}} & \textit{\textbf{.72}} & \textit{\textbf{.70}} & \textit{\textbf{.72}} & \textit{\textbf{.70}} & .68 \\
    Dense-2 & .56 & \textit{\textbf{.68}} & .64 & .65 & \textit{\textbf{.70}} & \textit{\textbf{.70}} & \textit{\textbf{.70}} & \textit{\textbf{.72}} & .67 & \textit{\textbf{.69}} \\
    Dense-3 & .53 & .67 & .62 & .65 & \textit{\textbf{.69}} & \textit{\textbf{.69}} & \textit{\textbf{.70}} & \textit{\textbf{.72}} & .67 & \textit{\textbf{.69}} \\
    Dense-4 & .55 & .68 & .62 & .64 & \textit{\textbf{.69}} & \textit{\textbf{.69}} & \textit{\textbf{.69}} & \textit{\textbf{.72}} & .66 & \textit{\textbf{.69}} \\
    \midrule
    Res-1 & .56 & .66 & .66 & .63 & \textit{\textbf{.72}} & \textit{\textbf{.72}} & \textit{\textbf{.71}} & \textit{\textbf{.72}} & \textit{\textbf{.70}} & .70 \\
    Res-2 & .55 & \textit{\textbf{.68}} & .64 & .60 & .67 & \textit{\textbf{.68}} & \textit{\textbf{.70}} & \textit{\textbf{.70}} & .67 & .65 \\
    Res-3 & .55 & \textbf{.69} & .64 & .60 & .67 & .68 & \textit{\textbf{.70}} & \textit{\textbf{.70}} & \textit{\textbf{.69}} & \textbf{.68} \\
    Res-4 & .55 & \textbf{.70} & .66 & .62 & \textbf{.71} & \textbf{.70} & \textit{\textbf{.72}} & .67 & \textbf{.71} & .68 \\
    \midrule
    Mean & .55 & \textbf{.65} & .62 & .59 & \textit{\textbf{.67}} & \textit{\textbf{.67}} & \textit{\textbf{.68}} & \textit{\textbf{.66}} & \textbf{.68} & \textbf{.66} \\
    Std dev & .02 & .05 & .04 & .05 & .04 & .04 & .04 & .06 & .02 & .05 \\
    \midrule
    Model & SFT & EWC & SST-1 & SST-B & JM & JST-1 & JST-B & & & 20k \\
    \midrule
    Conv-1 & \textit{\textbf{.58}} & \textit{\textbf{.56}} & .54 & --- & .51 & .51 & --- & & & .57 \\
    Conv-2 & \textit{\textbf{.63}} & .59 & .60 & --- & .61 & .58 & --- & & & .63 \\
    Conv-3 & \textit{\textbf{.64}} & .62 & .61 & --- & .62 & .58 & --- & & & .64 \\
    Conv-4 & \textit{\textbf{.64}} & \textit{\textbf{.62}} & .61 & --- & .61 & .59 & --- & & & .64 \\
    \midrule
    Dense-1 & .68 & .65 & .60 & .61 & \textit{\textbf{.71}} & .68 & .66 & & & .71 \\
    Dense-2 & .67 & .66 & .56 & .66 & \textit{\textbf{.70}} & .65 & .65 & & & .70 \\
    Dense-3 & .68 & .67 & .64 & .65 & .64 & .65 & .66 & & & .71 \\
    Dense-4 & \textit{\textbf{.68}} & .68 & .64 & .68 & .64 & .66 & .64 & & & .70 \\
    \midrule
    Res-1 & \textit{\textbf{.71}} & \textit{\textbf{.70}} & .66 & .65 & \textit{\textbf{.71}} & .67 & .66 & & & .72 \\
    Res-2 & \textit{\textbf{.68}} & .65 & .67 & .61 & .67 & .61 & .62 & & & .70 \\
    Res-3 & .67 & .64 & .67 & .65 & .68 & .64 & .62 & & & .71 \\
    Res-4 & .69 & .65 & .69 & .66 & .68 & .64 & .58 & & & .73 \\
    \midrule
    Mean & \textit{\textbf{.66}} & \textbf{.64} & .64 & .63 & \textbf{.65} & .65 & .61 & & & .68 \\
    Std dev & .03 & .04 & .04 & .04 & .05 & .02 & .04 & & & .05 \\
    \bottomrule
    \end{tabular}
    \caption[Test accuracy at the final time step for each model architecture for the shift sequence with 6k source samples, 4k samples with +30 red tints (T), 6k samples with +60 T, and 4k samples with +90 T.]{Test accuracy at the final time step for each model architecture for the shift sequence with 6k source samples, 4k samples with +30 red tints (T), 6k samples with +60 T, and 4k samples with +90 T.
    
    Oracle is last column (20k). Italicized if oracle-like (within 0.02 or better than oracle) for that architecture. Bolded if oracle-like or within 0.02 of best for that architecture (row). Mean is italicized if the average difference between the method and the oracle is less than 1 standard deviation below 0. Mean is bolded if the average difference between the method and either the best or the oracle is less than 1 standard deviation below 0.}
    \label{tab:ttt_results_full}
\end{table}

\begin{table}[htbp]
    \centering
    \begin{tabular}{lcccccccccc}
    \toprule
    Model & 4k & ERM & IRM & DRO & FT & LP-FT & I-FT & D-FT & ST-1 & ST-B \\
    \midrule
    Conv-1 & .37 & .37 & \textit{\textbf{.44}} & .39 & \textit{\textbf{.42}} & \textit{\textbf{.43}} & \textit{\textbf{.47}} & \textit{\textbf{.45}} & .40 & --- \\
    Conv-2 & .39 & .37 & .36 & .38 & .38 & .36 & .38 & .41 & .41 & --- \\
    Conv-3 & .40 & .34 & .36 & .36 & .40 & .39 & .37 & .36 & .37 & --- \\
    Conv-4 & .40 & .36 & .35 & .34 & .40 & .39 & .35 & .34 & .35 & --- \\
    \midrule
    Dense-1 & .43 & \textit{\textbf{.54}} & .52 & \textit{\textbf{.55}} & \textit{\textbf{.56}} & \textit{\textbf{.57}} & \textit{\textbf{.57}} & \textit{\textbf{.58}} & \textit{\textbf{.57}} & .53 \\
    Dense-2 & .41 & .54 & .54 & .55 & \textit{\textbf{.56}} & \textit{\textbf{.58}} & .55 & \textit{\textbf{.58}} & .54 & \textit{\textbf{.58}} \\
    Dense-3 & .42 & .53 & .53 & .54 & \textit{\textbf{.57}} & \textit{\textbf{.58}} & \textit{\textbf{.55}} & \textit{\textbf{.58}} & .54 & \textit{\textbf{.57}} \\
    Dense-4 & .40 & .53 & .54 & .54 & \textit{\textbf{.58}} & \textit{\textbf{.57}} & .53 & \textit{\textbf{.57}} & .49 & \textit{\textbf{.56}} \\
    \midrule
    Res-1 & .45 & .52 & .55 & .56 & \textit{\textbf{.59}} & \textit{\textbf{.60}} & \textit{\textbf{.59}} & \textit{\textbf{.59}} & .55 & .57 \\
    Res-2 & .39 & .54 & .53 & .54 & \textbf{.56} & \textbf{.56} & \textit{\textbf{.58}} & \textbf{.56} & .56 & .54 \\
    Res-3 & .38 & \textbf{.56} & .53 & .53 & \textbf{.56} & \textbf{.56} & \textbf{.56} & \textbf{.55} & \textit{\textbf{.57}} & .54 \\
    Res-4 & .35 & .53 & .54 & .49 & \textit{\textbf{.57}} & \textit{\textbf{.57}} & \textbf{.55} & .51 & \textit{\textbf{.57}} & .55 \\
    \midrule
    Mean & .40 & \textbf{.48} & \textbf{.48} & \textbf{.48} & \textit{\textbf{.51}} & \textit{\textbf{.51}} & \textit{\textbf{.50}} & \textit{\textbf{.51}} & \textbf{.55} & \textbf{.50} \\
    Std dev & .03 & .08 & .08 & .08 & .08 & .09 & .08 & .09 & .02 & .08 \\
    \midrule
    Model & SFT & EWC & SST-1 & SST-B & JM & JST-1 & JST-B & & & 20k \\
    \midrule
    Conv-1 & \textit{\textbf{.46}} & \textit{\textbf{.45}} & \textit{\textbf{.42}} & --- & .41 & .41 & --- & & & .43 \\
    Conv-2 & \textit{\textbf{.49}} & \textit{\textbf{.48}} & .45 & --- & .36 & .46 & --- & & & .49 \\
    Conv-3 & \textit{\textbf{.50}} & \textit{\textbf{.49}} & .45 & --- & .35 & .46 & --- & & & .50 \\
    Conv-4 & \textit{\textbf{.51}} & \textit{\textbf{.49}} & .46 & --- & .36 & .44 & --- & & & .49 \\
    \midrule
    Dense-1 & \textit{\textbf{.54}} & \textit{\textbf{.55}} & \textit{\textbf{.55}} & .53 & \textit{\textbf{.57}} & .52 & .51 & & & .56 \\
    Dense-2 & .56 & .55 & .50 & .54 & .48 & .50 & .49 & & & .58 \\
    Dense-3 & \textit{\textbf{.55}} & \textit{\textbf{.55}} & .51 & .53 & .47 & .50 & .49 & & & .56 \\
    Dense-4 & .54 & .54 & .51 & .52 & .49 & .49 & .51 & & & .57 \\
    \midrule
    Res-1 & .58 & \textit{\textbf{.60}} & .58 & .58 & .58 & .54 & .54 & & & .60 \\
    Res-2 & .54 & .50 & .55 & .52 & .54 & .53 & .53 & & & .59 \\
    Res-3 & .53 & .53 & .55 & .49 & .54 & .53 & .49 & & & .59 \\
    Res-4 & .53 & .50 & \textbf{.55} & .48 & .54 & .48 & .49 & & & .58 \\
    \midrule
    Mean & \textit{\textbf{.53}} & \textit{\textbf{.52}} & \textbf{.54} & \textbf{.50} & \textbf{.47} & .51 & .48 & & & .55 \\
    Std dev & .03 & .04 & .03 & .04 & .08 & .02 & .04 & & & .05 \\
    \bottomrule
    \end{tabular}
    \caption[Test accuracy at the final time step for each model architecture for the shift sequence with 6k source samples, 4k samples with 1 new sub-class (s), 6k samples with 2 new sub-classes, and 4k samples with 3 new sub-classes.]{Test accuracy at the final time step for each model architecture for the shift sequence with 6k source samples, 4k samples with 1 new sub-class (s), 6k samples with 2 new sub-classes, and 4k samples with 3 new sub-classes.
    
    Oracle is last column (20k). Italicized if oracle-like (within 0.02 or better than oracle) for that architecture. Bolded if oracle-like or within 0.02 of best for that architecture (row). Mean is italicized if the average difference between the method and the oracle is less than 1 standard deviation below 0. Mean is bolded if the average difference between the method and either the best or the oracle is less than 1 standard deviation below 0.}
    \label{tab:sss_results_full}
\end{table}

\begin{table}[htbp]
    \centering
    \begin{tabular}{lcccccccccc}
    \toprule
    Model & Final & ERM & IRM & DRO & FT & LP-FT & I-FT & D-FT & ST-1 & ST-B \\
    \midrule
    Conv-1 & .61 & .88 & .87 & .83 & .90 & .92 & .75 & \textbf{.94} & .83 & --- \\
    Conv-2 & .84 & .90 & .92 & .86 & .94 & .93 & .90 & .91 & .93 & --- \\
    Conv-3 & \textbf{.95} & .92 & \textbf{.95} & .92 & \textbf{.95} & \textbf{.95} & \textbf{.94} & \textbf{.96} & .93 & --- \\
    Conv-4 & .91 & .94 & \textbf{.95} & \textbf{.94} & \textbf{.95} & \textbf{.96} & \textbf{.95} & \textbf{.94} & \textbf{.95} & --- \\
    \midrule
    Dense-1 & .73 & .87 & \textbf{.94} & .91 & .93 & \textbf{.95} & \textbf{.94} & .93 & .85 & \textbf{.94} \\
    Dense-2 & .89 & .89 & \textbf{.95} & .92 & .87 & .93 & .93 & .92 & .88 & \textbf{.94} \\
    Dense-3 & .89 & .87 & \textbf{.94} & .93 & .91 & .91 & .93 & .92 & .84 & .81 \\
    Dense-4 & .72 & .87 & \textbf{.93} & .91 & .91 & .91 & \textbf{.92} & .81 & .90 & .86 \\
    \midrule
    Res-1 & .86 & .86 & .92 & .87 & \textbf{.94} & .94 & .92 & .93 & .90 & .90 \\
    Res-2 & .82 & .90 & \textbf{.96} & .94 & .93 & .92 & \textbf{.97} & \textbf{.96} & .90 & .94 \\
    Res-3 & .90 & .53 & \textbf{.94} & .89 & \textbf{.93} & .92 & \textbf{.93} & .90 & \textbf{.94} & .92 \\
    Res-4 & .73 & .86 & .92 & .78 & .51 & .85 & \textbf{.94} & .79 & .82 & .80 \\
    \midrule
    Mean & .82 & \textbf{.86} & \textbf{.93} & .89 & \textbf{.89} & .93 & \textbf{.92} & \textbf{.91} & .88 & .90 \\
    Std dev & .10 & .10 & .02 & .05 & .12 & .03 & .05 & .05 & .04 & .05 \\
    \midrule
    Model & SFT & EWC & SST-1 & SST-B & JM & JST-1 & JST-B & & & \\
    \midrule
    Conv-1 & \textbf{.93} & .91 & .85 & --- & .60 & .51 & --- \\
    Conv-2 & \textbf{.96} & \textbf{.96} & .94 & --- & .51 & .51 & --- \\
    Conv-3 & \textbf{.95} & \textbf{.96} & .93 & --- & .51 & .85 & --- \\
    Conv-4 & \textbf{.96} & \textbf{.95} & \textbf{.95} & --- & .51 & .84 & --- \\
    \midrule
    Dense-1 & \textbf{.94} & \textbf{.94} & .67 & .89 & .82 & .87 & .87 \\
    Dense-2 & \textbf{.95} & \textbf{.95} & .92 & .91 & .86 & .87 & .85 \\
    Dense-3 & \textbf{.96} & .92 & .91 & .88 & .85 & .51 & .84 \\
    Dense-4 & \textbf{.94} & .86 & \textbf{.94} & .81 & .80 & .51 & .52 \\
    \midrule
    Res-1 & \textbf{.96} & \textbf{.96} & .88 & .91 & .80 & .89 & .51 \\
    Res-2 & \textbf{.96} & .95 & .93 & .90 & .53 & .51 & .88 \\
    Res-3 & \textbf{.93} & .89 & .91 & .78 & .51 & .51 & .51 \\
    Res-4 & .89 & .91 & .90 & .72 & .51 & .59 & .51 \\
    \midrule
    Mean & \textbf{.94} & \textbf{.93} & \textbf{.88} & .87 & .65 & .66 & .68 \\
    Std dev & .02 & .03 & .08 & .07 & .15 & .17 & .17 \\
    \bottomrule
    \end{tabular}
    \caption[Test accuracy on 2000s portraits for each model architecture for the Portraits sequence.]{Test accuracy on 2000s portraits for each model architecture for the Portraits sequence.
    
    Oracle is last column (20k). Italicized if oracle-like (within 0.02 or better than oracle) for that architecture. Bolded if oracle-like or within 0.02 of best for that architecture (row). Mean is italicized if the average difference between the method and the oracle is less than 1 standard deviation below 0. Mean is bolded if the average difference between the method and either the best or the oracle is less than 1 standard deviation below 0.}
    \label{tab:portraits_results_full}
\end{table}

\begin{table}[htbp]
    \centering
    \begin{tabular}{ccccccccccc}
    \toprule
    \# steps & 4k & ERM & IRM & DRO & FT & LP-FT & I-FT & D-FT & ST-1 & ST-B \\
    \midrule
    2 & .55 & .62 & .61 & .59 & .62 & .62 & .61 & .62 & .62 & .62 \\
    3 & .53 & .58 & \textbf{.59} & .58 & \textbf{.60} & .58 & \textbf{.61} & .59 & \textbf{.59} & .58 \\
    4 & .56 & .59 & .58 & .57 & .58 & .59 & \textbf{.62} & \textbf{.61} & \textbf{.62} & \textbf{.62} \\
    5 & .53 & .55 & .56 & .49 & .57 & \textbf{.59} & \textbf{.60} & \textbf{.60} & \textbf{.58} & .58 \\
    6 & .53 & .55 & .52 & .49 & \textbf{.57} & \textbf{.58} & \textbf{.59} & \textbf{.58} & .57 & \textbf{.58} \\
    7 & .56 & .54 & .55 & .50 & \textbf{.60} & \textbf{.60} & \textbf{.62} & \textbf{.60} & \textbf{.60} & \textbf{.60} \\
    8 & .53 & .54 & .50 & .47 & \textbf{.57} & \textbf{.57} & \textbf{.58} & \textbf{.58} & .56 & \textbf{.57} \\
    \midrule
    Mean & .54 & .57 & .56 & .53 & .59 & .59 & \textbf{.60} & \textbf{.60} & \textbf{.59} & \textbf{.59} \\
    Std dev & .01 & .03 & .04 & .05 & .02 & .01 & .01 & .01 & .02 & .02 \\
    \midrule
    \# steps & SFT & EWC & SST-1 & SST-B & JM & JST-1 & JST-B & & & 20k \\
    \midrule
    2 & \textbf{.63} & \textbf{.65} & .58 & .57 & .58 & .54 & .57 & & & .71 \\
    3 & \textbf{.61} & \textbf{.59} & .58 & .57 & \textbf{.59} & .57 & .53 & & & .69 \\
    4 & .59 & .60 & \textbf{.62} & \textbf{.61} & \textbf{.61} & .59 & .55 & & & .74 \\
    5 & \textbf{.59} & .56 & \textbf{.60} & .56 & .57 & .58 & .47 & & & .69 \\
    6 & .55 & .55 & \textbf{.57} & .56 & \textbf{.57} & \textbf{.58} & .48 & & & .70 \\
    7 & .58 & .56 & .58 & \textbf{.60} & \textbf{.60} & \textbf{.61} & .43 & & & .71 \\
    8 & .56 & .55 & .55 & .55 & \textbf{.58} & \textbf{.57} & .41 & & & .69 \\
    \midrule
    Mean & \textbf{.59} & \textbf{.58} & .58 & .57 & .59 & .58 & .49 & & & .70 \\
    Std dev & .02 & .03 & .02 & .02 & .01 & .02 & .06 & & & .02 \\
    \bottomrule
    \end{tabular}
    \caption[Test accuracy at final time step for sequence of rotations on CIFAR-10 data with 4-block ResNet architecture.]{Test accuracy at final time step for sequence of rotations on CIFAR-10 data with 4-block ResNet architecture.\\
    
    Oracle is last column (20k). Italicized if oracle-like (within 0.02 or better than oracle) for that architecture. Bolded if oracle-like or within 0.02 of best for that architecture (row). Mean is italicized if the average difference between the method and the oracle is less than 1 standard deviation below 0. Mean is bolded if the average difference between the method and either the best or the oracle is less than 1 standard deviation below 0.}
    \label{tab:rotation_seq_results}
\end{table}

\clearpage
\section{Additional Analyses of the Joint Model}
\label{app:joint_model_analysis}

To our knowledge, the joint modeling approach we introduce in Section~\ref{sec:methods_leverage_seq} is a novel method. In this appendix, we examine the two mechanisms for building sequential structure into the joint model: 1) regularization on the difference between weights at adjacent time steps and 2) adding new side modules at each step parallel to the side modules from previous steps. We also include an experiment that directly compares different aspects of the joint model and sequential fine-tuning.

The joint model with separate layers at each time step minimizes a weighted sum of the losses across all time steps. In addition to the standard L2 regularization on the model weights, there is an additional regularization term on the difference between weights at adjacent time steps. The training objective is
\begin{equation}
    \sum_{t=0}^T c_t \mathcal{L}\left(\mathcal{M}_t, \mathcal{D}_t\right) + \lambda_2 \sum_{t=0}^T \lVert \mathbf{w}_t \rVert_2^2 + \lambda_{adj} \sum_{t=1}^T \lVert \mathbf{w}_{t} - \mathbf{w}_{t-1} \rVert_2^2
\end{equation}
where $c_t$ is the weight on the loss at time step $t$, $\mathcal{M}_t$ is the joint model with the modules for time $t$ activated for predictions, $\lambda_2$ is the standard L2 regularization constant, and $\lambda_{adj}$ is the regularization constant on the difference between weights at adjacent time steps. We set $c_T = 3$ and $c_t = 1$ for $0 \le t < T$.

\begin{figure}[htbp]
    \centering
    \begin{minipage}[t]{.65\textwidth}
        \centering
        \includegraphics[width=\textwidth]{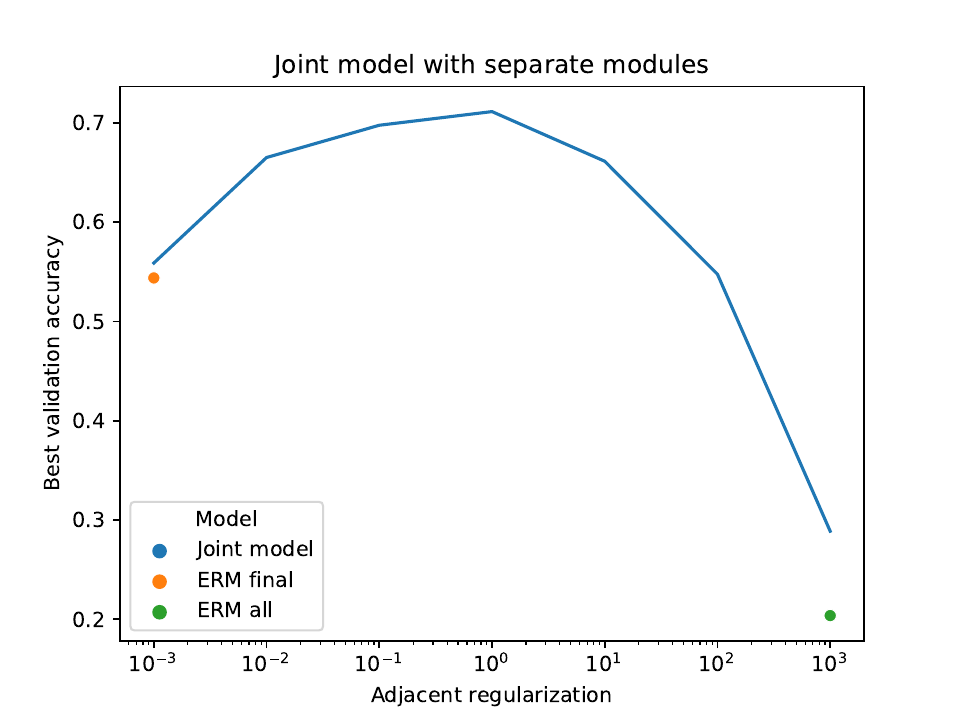}
    \end{minipage}
    \caption[Validation accuracy at the final time step vs regularization on the difference between parameters at adjacent time steps in a joint model.]{Validation accuracy at the final time step vs regularization on the difference between parameters at adjacent time steps in a joint model. This was from the joint model with 4 residual blocks trained on the sequence with rotations, corruptions, and then label flips. No adjacent regularization is plotted at 0.001 on the $x$-axis.}
    \label{fig:adjacent_reg_val_acc}
\end{figure}

\textbf{Joint model benefits from regularization bringing weights at adjacent time steps closer together} We can examine the effect of the regularization on the difference between weights at adjacent time steps by varying $\lambda_{adj}$ between two extremes: 1) When there is no adjacent regularization ($\lambda_{adj} = 0$), all time steps can have completely different models. This means the earlier data is not leveraged, so the joint model at the final time step would perform the same as ERM on data from just the final time step (ERM final). 2) When there is very high adjacent regularization ($\lambda_{adj} \rightarrow \infty$), all the time steps are forced to have the same parameters, so this would be equivalent to a single model learned with ERM on data from all distributions (ERM all). Figure~\ref{fig:adjacent_reg_val_acc} shows the validation accuracy of the joint model indeed traverses from the validation accuracy of ERM final to ERM all. Between the two extremes, the validation accuracy is higher at intermediate levels of adjacent regularization. This means the joint model benefits from leveraging data at historical time points when there are mild constraints on parameters at adjacent time steps being close together.

\textbf{Side-tuning is constrained by additive output structure} In Section~\ref{sec:methods_leverage_seq}, we defined the variant of the joint model with side modules to output $B\left(x\right) + \sum_{j=1}^t S_j\left(x\right)$ at time $t$. A similar idea was introduced for fine-tuning and sequential fine-tuning. In all 3 cases, the goal of introducing a small side module is to improve parameter efficiency. However, the trade-off is a 1-layer side module for each block may not have enough capacity to model the change at each time step. Indeed, from Tables~\ref{tab:clr_results_full}-\ref{tab:rotation_seq_results}, we see that the performance from fitting a 1-layer side module at each time step (ST-1, SST-1, and JST-1) is often worse than fine-tuning the entire module (FT and SFT) or fitting a separate module at each time step (JM). This deterioration is not just a result of having fewer parameters. When we incorporate side modules that are the same size as the original modules (ST-B, SST-B, and JST-B), the performance is often still not as good as the equivalent fine-tuning or joint model variant. Even though a side module that is the same size as the original block has the same number of parameters, the model is more constrained because linearly adding the outputs at the end of each block does not allow for as much flexibility as changing the parameters within each block. Adding side modules does not have the same capacity as updating each block even when the side modules have the same number of parameters. This raises some questions about whether side modules are truly parameter-efficient as the parameters may not be used as efficiently.

\textbf{Initialization from previous steps that is essential in sequential fine-tuning is lacking in joint model} As discussed in the Section~\ref{sec:results}, the joint model did not seem to benefit from accessing data from all steps simultaneously since it did not outperform sequential fine-tuning. To understand why, we considered the effect of initialization versus regularization towards previous parameters. Fine-tuning initializes the model at time $t$ with parameters from time $t - 1$, while the joint model initializes all modules randomly and applies regularization on the difference between adjacent parameters to bring them closer. Does the initialization at previous parameters provider a better starting point for optimization? To answer this question, we ablated the initialization during fine-tuning and examined whether regularization towards previous parameters during fine-tuning is able to recover the same benefits as initialization. Table~\ref{table:erm_fine_tune_v_frozen_joint} shows how fine-tuning with the initialization ablated and regularization towards previous parameters is able to get close to but not quite match standard fine-tuning. This highlights how initialization, which is lacking in the joint model, might be important for learning a good model at the final time point.

As a sanity check, Table~\ref{table:erm_fine_tune_v_frozen_joint} also shows how the joint model implementation can replicate equivalent settings from fine-tuning. These ablations are implemented with a 2-step joint model, where $t = 0$ is frozen at the model fit on all previous time steps and $t = 1$ is fit on data from the final time-step with the initialization and regularization set to match the fine-tuning variations. The joint model with the prior step frozen has almost equivalent performance when compared to the corresponding fine-tuning variations.

\begin{table}[htb]
    \centering
    \begin{tabular}{lc}
    \toprule
    Method & Test accuracy \\
    \midrule
    ERM final & .5470 \\
    Standard fine-tuning & .6766 \\
    Fine-tuning w/ reg towards previous & .6726 \\
    Fine-tuning w/ init ablated + reg to prev & .6644 \\
    \midrule
    Frozen joint w/o prev init or reg to prev & .5466 \\
    Frozen joint w/ prev init & .6784 \\
    Frozen joint w/ prev init + reg to prev & .6756 \\
    Frozen joint w/ only reg to prev & .6564 \\
    \bottomrule
    \end{tabular}
    \caption[Test accuracy from experiments examining the relative importance of initialization and regularization towards previous models]{Test accuracy from variants of fine-tuning that examine the importance of initialization at previous parameters versus regularization towards previous parameters. This experiment was performed on the rotation, corruption, then label flip shift sequence with the Res-4 architecture. The frozen joint model has the $t = 0$ modules frozen at the parameters from a single Res-4 model fit on data from all previous time points. Thus, only the $t = 1$ modules are fit on data from the final time step. That makes each row $i + 4$ equivalent to row $i$ for $i = 0, 1, 2, 3$. The matching performances provide assurance that the joint model is implemented correctly.}
    \label{table:erm_fine_tune_v_frozen_joint}
\end{table}

\clearpage
\section{Additional Visualizations}

In this appendix, we discuss motivation and related work for the linear interpolation visualization created in Section~\ref{sec:lin_interpolate_visual}. We also present the visuals for other synthetic sequences alongside additional observations. Then, we provide preliminary groundwork for visualizing sequences of models over multiple time steps.

\subsection{Additional Analysis of Linear Interpolation Visuals}
\label{app:lin_interpolation_visual}

In Section~\ref{sec:lin_interpolate_visual}, we define the linear interpolation path by linearly interpolating the weights from the initialization to the final model at the last time step and plotting the test accuracy at the last time step from evaluating the model with interpolated weights. That is, the interpolated model $\mathcal{M}_s$ is defined by weights $\left(1 - s\right) \mathbf{w}_{T-1} + s \mathbf{w}_T$ for $0 \le s \le 1$. The $x$-coordinate is $s$, and the $y$-coordinate is $\mathcal{L}\left(\mathcal{M}_s, \mathcal{D}_T\right)$. We visualize the linear interpolation path because it is more informative about the loss landscape than the update path taken by gradient descent. Gradient descent may take a very circumspect path towards the final optimum. While visualizing the loss landscape around this path can also reflect whether the initialization and the final model are separated by a loss barrier, such a plot is not feasible in two dimensions. The weights are very high-dimensional, so the loss landscape---a function of those high-dimensional weights---cannot be visualized easily.

\citet{neyshabur2020being} and \citet{mehta2023empirical} also visualize linear interpolation paths in their work. \citet{neyshabur2020being} plot the test accuracy when linearly interpolating the weights between two randomly initialized models or between two fine-tuned models. They show the accuracy is fairly constant between two fine-tuned models while there is an accuracy basin between the two randomly initialized models. They infer from the accuracy basin that the two randomly initialized models are separated by a performance barrier. \citet{mehta2023empirical} use linear interpolation to examine continual learning. They plot the loss on the first task when linearly interpolating the model weights between the weights learned for the first and second task. They show that the loss does not increase as much for models initialized by pre-training, so pre-training helps with mitigating catastrophic forgetting. 

\begin{figure}[htbp]
    \centering
    \begin{minipage}[t]{.96\textwidth}
        \centering
        \includegraphics[width=\textwidth,trim={0 50 0 50},clip]{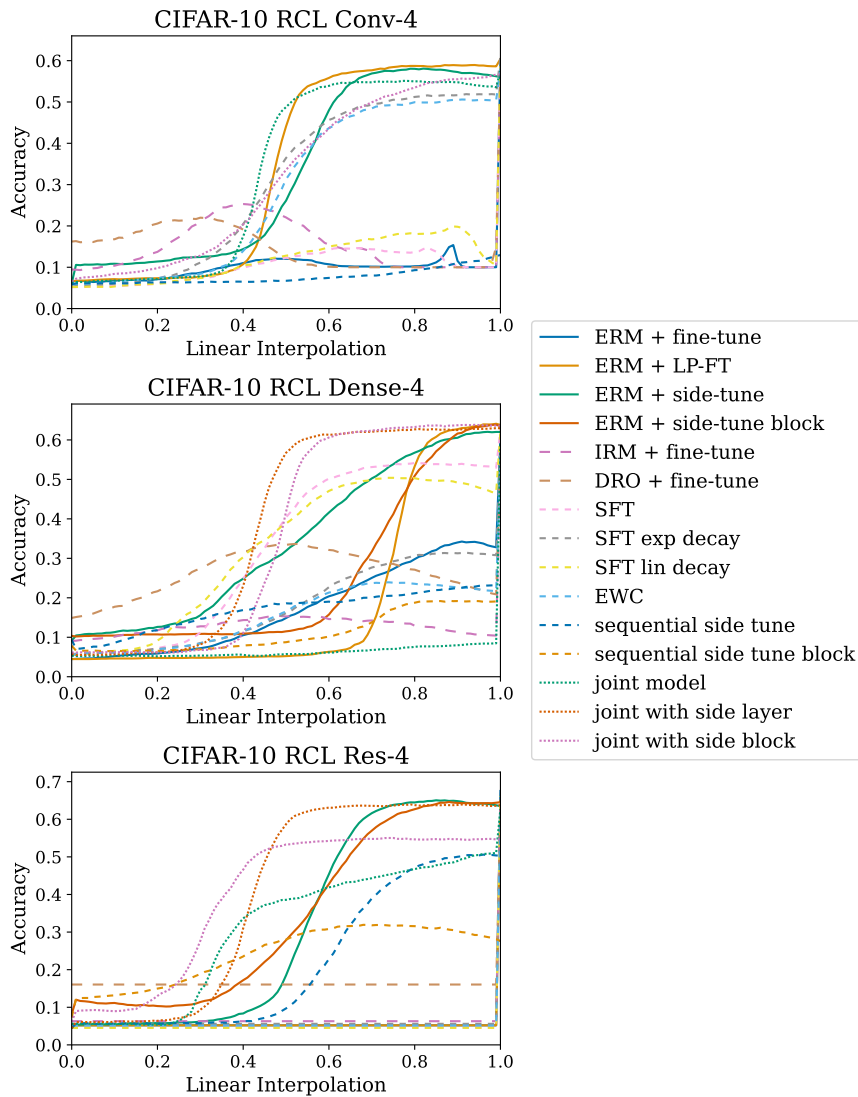}
    \end{minipage}
    \caption{Change in accuracy when linearly interpolating model weights from penultimate to final time step in the sequence with rotations, corruptions, and label flips.}
    \label{fig:rcl_lin_interpolate}
\end{figure}

\begin{figure}[htbp]
    \centering
    \begin{minipage}[t]{.96\textwidth}
        \centering
        \includegraphics[width=\textwidth,trim={0 50 0 50},clip]{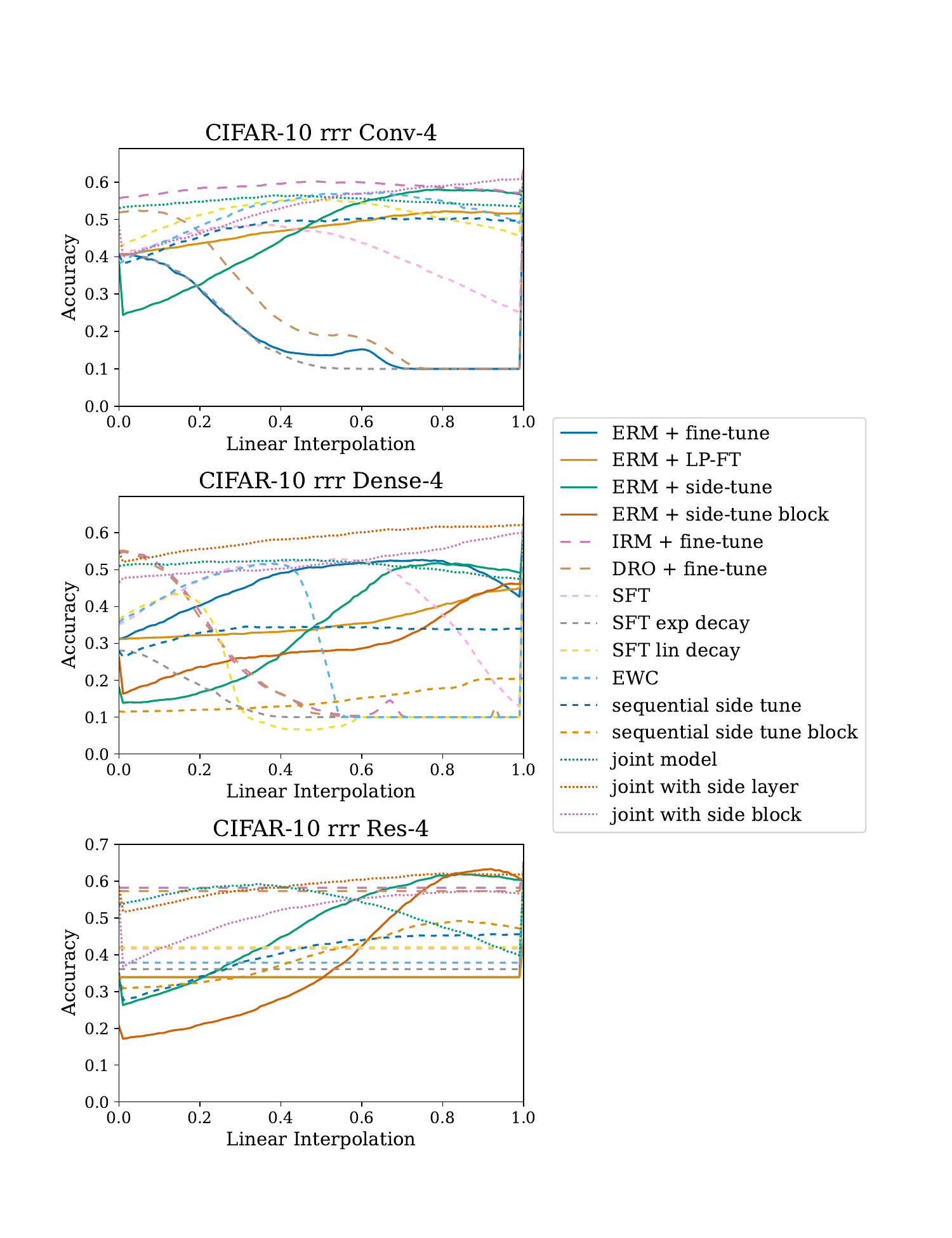}
    \end{minipage}
    \caption{Change in accuracy when linearly interpolating model weights from penultimate to final time step in the sequence with 3 conditional rotations.}
    \label{fig:rrr_lin_interpolate}
\end{figure}

\begin{figure}[htbp]
    \centering
    \begin{minipage}[t]{.96\textwidth}
        \centering
        \includegraphics[width=\textwidth,trim={0 50 0 50},clip]{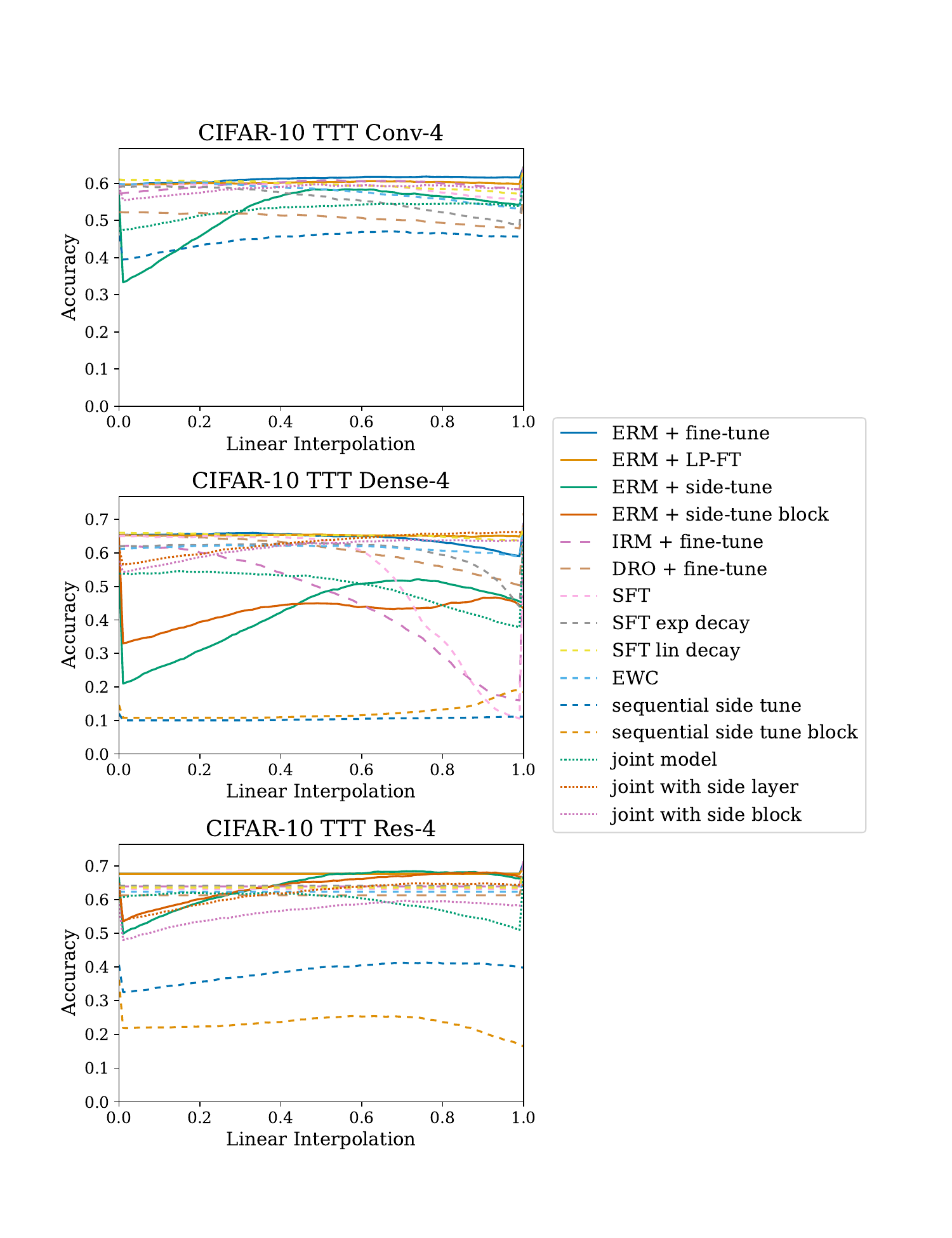}
    \end{minipage}
    \caption{Change in accuracy when linearly interpolating model weights from penultimate to final time step in the sequence with 3 applications of red tinting.}
    \label{fig:ttt_lin_interpolate}
\end{figure}

\begin{figure}[htbp]
    \centering
    \begin{minipage}[t]{.96\textwidth}
        \centering
        \includegraphics[width=\textwidth,trim={0 50 0 50},clip]{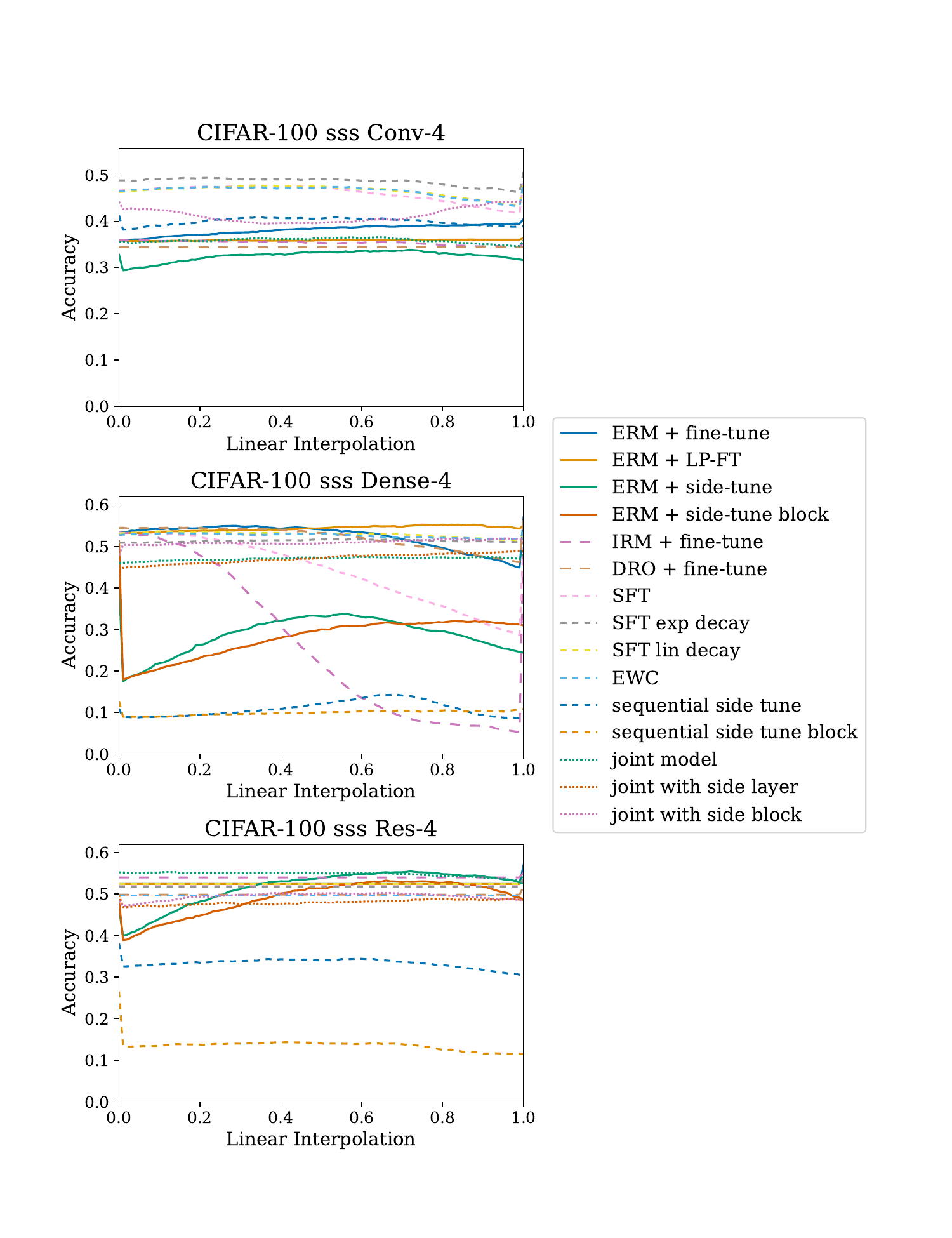}
    \end{minipage}
    \caption{Change in accuracy when linearly interpolating model weights from penultimate to final time step in the sequence with 3 sub-population shifts.}
    \label{fig:sss_lin_interpolate}
\end{figure}

Figures~\ref{fig:rcl_lin_interpolate}-\ref{fig:sss_lin_interpolate} visualize the linear interpolation paths for the 4 synthetic sequences that are not shown completely in Figure~\ref{fig:clr_lin_interpolate}. We can make the following observations:

\textbf{Side-tuning encourages a smooth linear interpolation path} For almost all sequences and architectures, starting with ERM on all historical data and then side-tuning either with a single layer or a single block results in a fairly smooth linear interpolation path. There may be a sharp drop in accuracy close to the initialization, a sharp increase close to the final optimum, and sometimes a gradual drop just before the sharp increase, but in general, the path increases smoothly. This suggests the additive side module structure may be a good way to constrain model updates to a smooth part of the loss landscape.

\textbf{Residual networks characterized by flat landscape throughout path and sharp local minimum} For a majority of the methods and for all sequences that do not end with label flips, the loss landscape is almost entirely flat other than a sharp drop near the initialization and a sharp increase near the final optimum. This suggests that the loss landscape for residual networks may be highly non-convex near the initialization and the optimum, and the loss plateau in the middle may be fairly wide. To understand just how wide the loss plateau is, it may be helpful to quantify the total change in weights along the linear interpolation path. 

\textbf{Smooth sigmoid shape for best-performing methods is observed for all architectures when final step is label flip} The trend observed for convolutional networks in Section~\ref{sec:lin_interpolate_visual} also holds for dense and residual networks. It seems like the smooth change in the output layer where the predictions flip occurs regardless of the feature representation in the preceding layers.

\textbf{Flat loss from initialization to final for sequences with little shift} For the sequence of red recolorings and the sequence of sub-population shifts, Figures~\ref{fig:ttt_lin_interpolate} and~\ref{fig:sss_lin_interpolate} show flat linear interpolation paths for many methods. This flatness suggests there may be little shift in these sequences, so the model learned on historical data already performs quite well at the last step. This conclusion is in agreement with the shift that is quantified in Appendix~\ref{app:quantifying_shift} and the performance metrics in Tables~\ref{tab:mean_results},~\ref{tab:ttt_results_full}, and~\ref{tab:sss_results_full}.

\begin{figure}[htbp]
    \centering
    \begin{minipage}[t]{.8\textwidth}
        \centering
        \includegraphics[width=\textwidth]{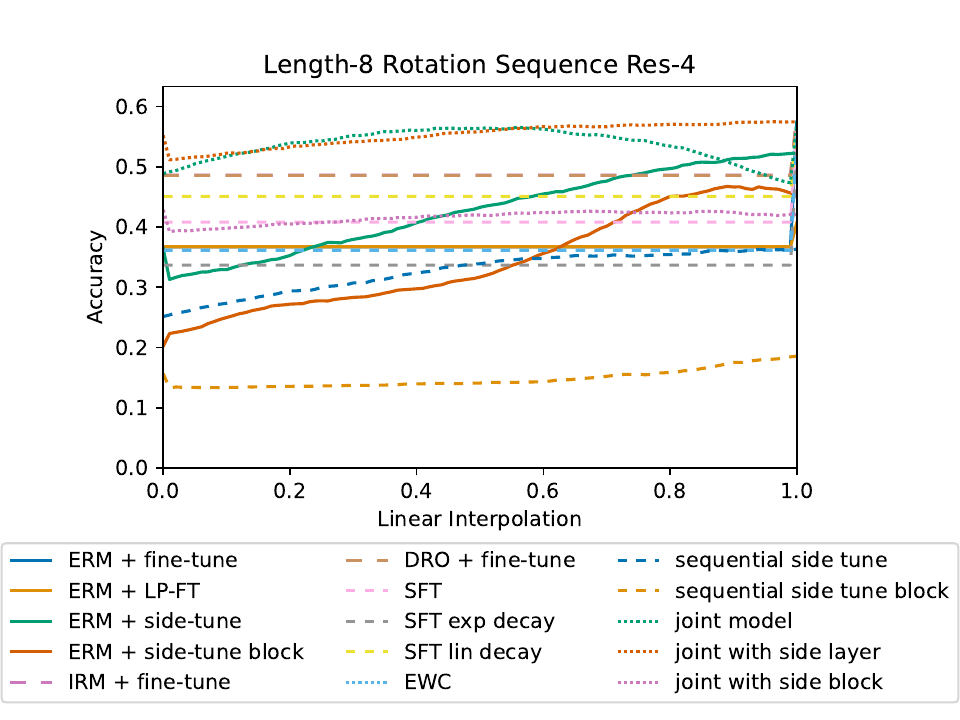}
    \end{minipage}
    \caption{Change in accuracy when linearly interpolating model weights from penultimate to final time step in the sequence with 7 conditional rotations.}
    \label{fig:rotation_seq_lin_interpolate}
\end{figure}

\begin{figure}[htbp]
    \centering
    \begin{minipage}[t]{.96\textwidth}
        \centering
        \includegraphics[width=\textwidth,trim={0 50 0 50},clip]{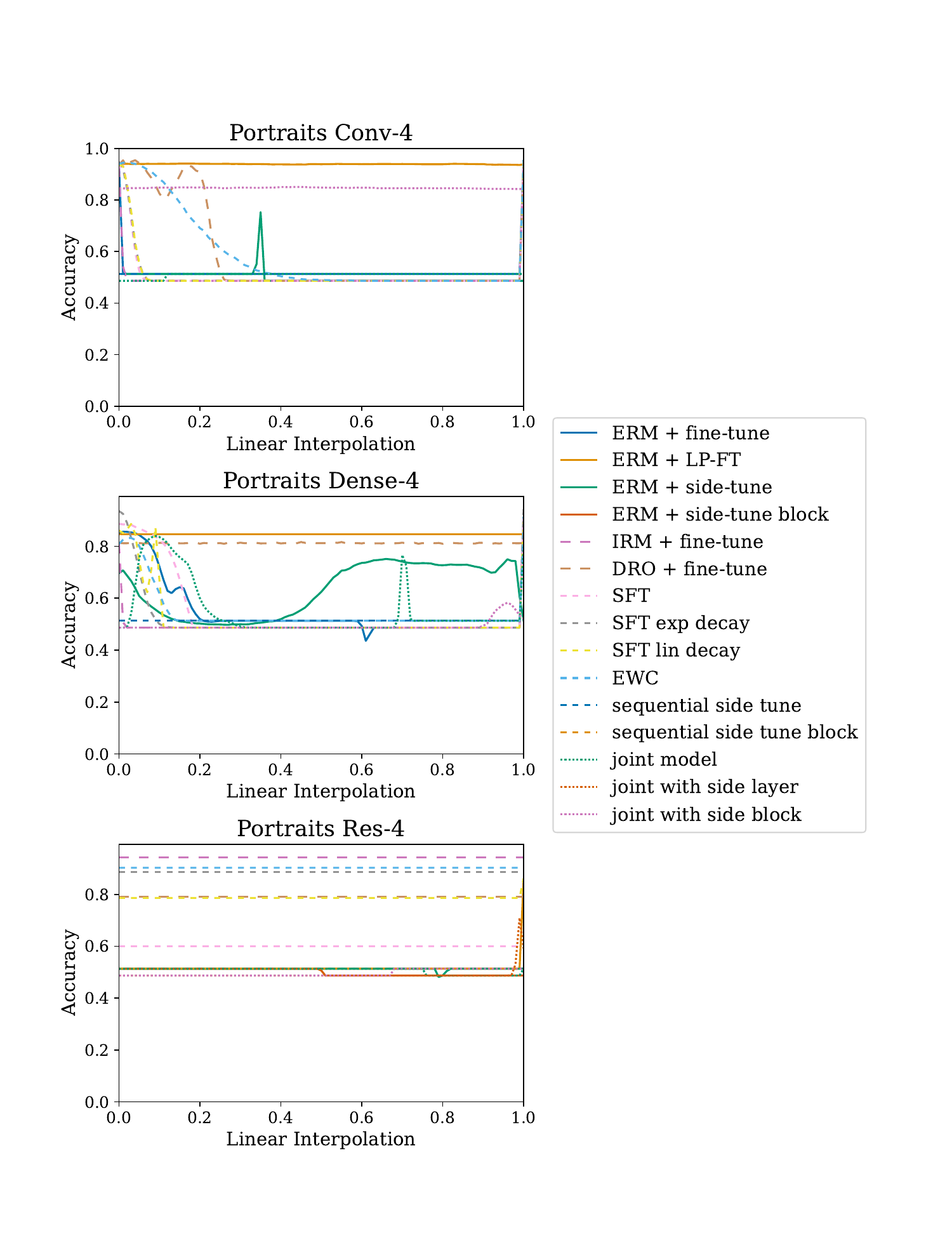}
    \end{minipage}
    \caption{Change in accuracy when linearly interpolating model weights from penultimate to final time step in the real-world Portraits sequence.}
    \label{fig:portraits_lin_interpolate}
\end{figure}

\textbf{Different trends are observed for the real-world Portraits dataset} The linear interpolation paths at the final step of the Portraits sequence in Figure~\ref{fig:portraits_lin_interpolate} look different from the plots for the synthetic sequences. For example, the joint model often does not exhibit the same smooth trend with little change. On the other hand, ERM followed by linear probing then fine-tuning on the final step has a completely flat path for both the convolutional and dense networks. Initializing the model via DRO on historical data followed by fine-tuning also produces a fairly flat interpolation path for the dense architecture. Many methods have a large loss plateau where the test accuracy stays flat at 0.5. Across this plateau, the interpolated models are essentially random since they are performing binary classification. The only pattern observed for synthetic sequences that also seems to hold for real-world sequences is the flat landscape observed for many methods with the residual network architecture. IRM followed by fine-tuning at the final step and a couple variants of sequential fine-tuning have paths that stay flat at a relatively high accuracy, suggesting that those methods are able to find a good initialization where minimal updates are needed. Finally, we note that this difference in trends for the real-world sequence is not attributed to the longer sequence length. As a comparison, we also plot the interpolation paths at the final step of the 8-step rotation sequence in Figure~\ref{fig:rotation_seq_lin_interpolate}. Here, the patterns regarding the joint model and ERM followed by side-tuning that were observed for length-4 sequences are preserved. Thus, the difference in trends suggests that the relationship between the initialization and the final optimum may not be the same for real-world shifts or binary classification tasks.

\clearpage
\subsection{Visualization Projecting Models over Time onto Two-Dimensional Space}

\label{app:visual_weight_proj_cca}

Because the sequential fine-tuning and joint model approaches learn a separate function at each time step, we are also interested in visualizing whether the models learned over the time steps gradually approach the final model. We may also be interested in directly comparing the models rather than the losses from different approaches---that is, how close are the weights or the intermediate outputs from the initial and final models produced by each method to the oracle learned solely on the final distribution? To analyze the evolution of models over time and visualize the model weights and behavior directly, we explore two additional visualization approaches: 1) Projecting the model weights onto two directions that reflect the shift from historical distributions to the final time point and the limited sample size at the final time point. 2) Using canonical correlation analysis to compute the top coefficients explaining most of the variance in the outputs from intermediate layers.

\textbf{Projection of model weights onto directions of historical shift and limited final sample size} Past works have focused on examining three models by placing the first model weight $\mathbf{w}_1$ at the origin and orienting the $x$-axis and $y$-axis such that they point from the first model weight to the second and third model weights $\mathbf{w}_2$ and $\mathbf{w}_3$, respectively. This approach was first used by \citet{garipov2018loss}. To define the axes, they let
\begin{align}
    \mathbf{u} &= \mathbf{w}_2 - \mathbf{w}_1 \\
    \mathbf{v} &= \mathbf{w}_3 - \mathbf{w}_1 - \langle \mathbf{w}_3 - \mathbf{w}_1, \mathbf{w}_2 - \mathbf{w}_1 \rangle \frac{\mathbf{w}_2 - \mathbf{w}_1}{\lVert \mathbf{w}_2 - \mathbf{w}_1 \rVert^2}
\end{align}
The axes are then defined as $\frac{\mathbf{u}}{\lVert \mathbf{u} \rVert}$ and $\frac{\mathbf{v}}{\lVert \mathbf{v} \rVert}$. Subtracting the projection onto the first axis when computing the second axis makes the two axes orthogonal. \citet{wortsman2022model} also use this approach to visualize the space between two fine-tuned models, and \citet{mehta2023empirical} use this approach to study catastrophic forgetting during continual learning across 3 tasks.

Since we are analyzing multiple models, we cannot just explore the space of affine combinations of these axes. Instead, we are projecting multiple models onto a 2-dimensional space. To do so, we also select three models to define the origin and projection directions. We want to capture the effects of distribution shift and limited sample size at the final time point. Thus, we define the origin by the weights of the final oracle model. Let these be $\mathbf{w}^*$. To represent how much a model $\mathbf{w}$ is affected by using historical shifted data, we define the $x$-axis as the projection of $\mathbf{w} - \mathbf{w}^*$ onto $\mathbf{w}_H - \mathbf{w}^*$, where $\mathbf{w}_H$ are the weights of a model trained with ERM on data from all previous time steps. To represent how much a model $\mathbf{w}$ is affected by limited data at the final time step, the $y$-axis is defined as length of the projection of $\mathbf{w} - \mathbf{w}^*$ onto $\mathbf{w}_T - \mathbf{w}^*$, where $\mathbf{w}_T$ are the weights of the model trained on limited data from the final time step. That is, the coordinates for a model $\mathbf{w}$ are defined as
\begin{align}
    x &= \langle \mathbf{w} - \mathbf{w}^*, \mathbf{w}_H - \mathbf{w}^* \rangle \frac{\mathbf{w}_H - \mathbf{w}^*}{\lVert \mathbf{w}_H - \mathbf{w}^* \rVert^2} \\
    y &= \langle \mathbf{w} - \mathbf{w}^*, \mathbf{w}_T - \mathbf{w}^* \rangle \frac{\mathbf{w}_T - \mathbf{w}^*}{\lVert \mathbf{w}_T - \mathbf{w}^* \rVert^2}
\end{align}
To preserve the meaning of these directions, we do not make the axes orthogonal by subtracting the projection of one difference onto the other axis.

\begin{figure}[htbp]
    \centering
    \begin{minipage}[t]{.96\textwidth}
        \centering
        \includegraphics[width=\textwidth,trim={60 0 60 15},clip]{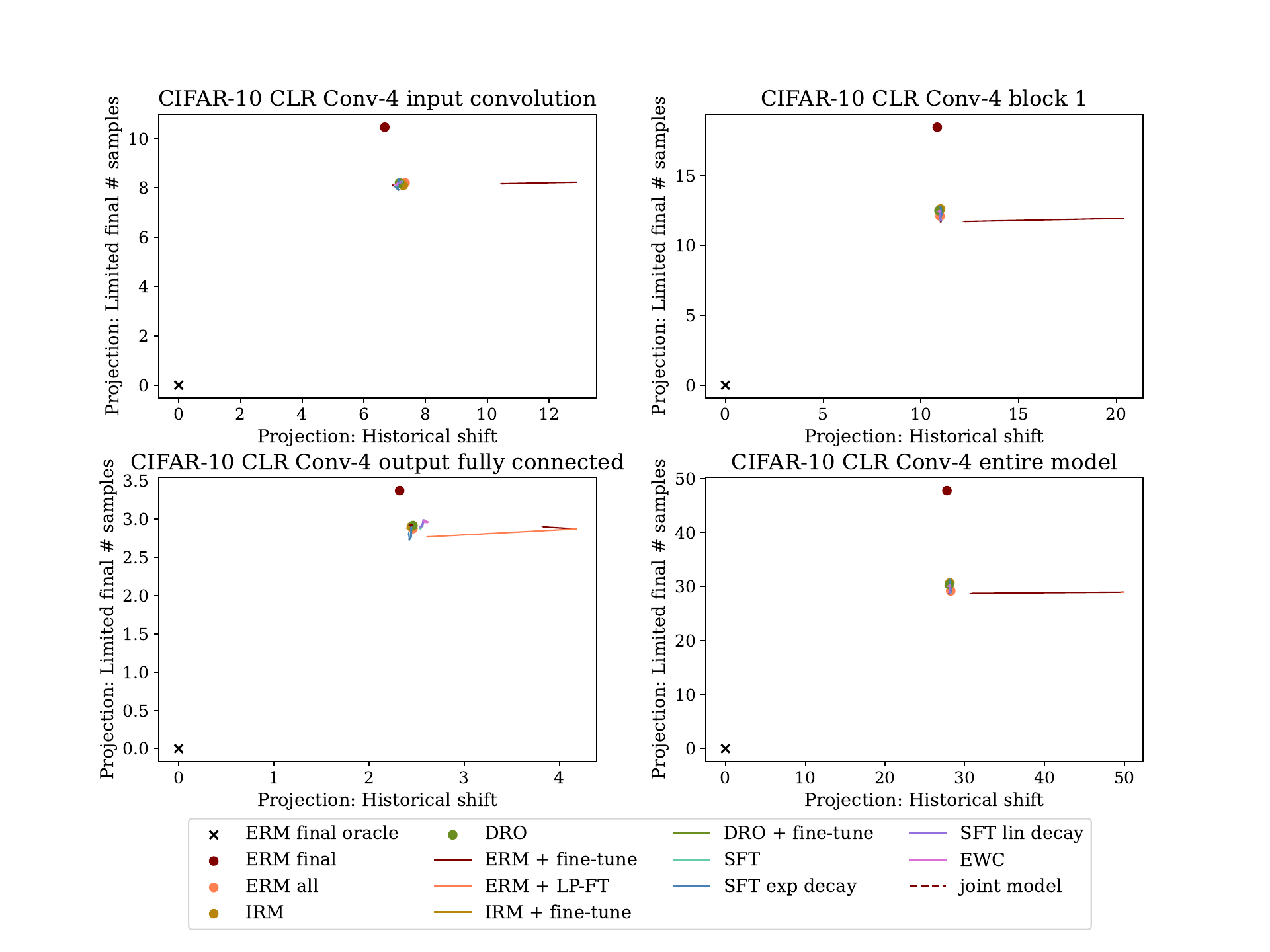}
    \end{minipage}
    \caption[Visualization of Conv-4 models over time for the corruption, label, and rotation shift sequence created by projecting the weights onto two directions that reflect historical shift and limited sample size at the final time step.]{Visualization of Conv-4 models over time for the corruption, label, and rotation (CLR) shift sequence created by projecting the weights onto a direction that reflects historical shift ($x$-axis) and a direction the reflects limited sample size at the final time step ($y$-axis). Models closer to $\left(0, 0\right)$ have weights that are more similar to the oracle. The subplot for the entire model concatenates the weights from all layers when computing the projection.}
    \label{fig:weight_proj_clr_visual}
\end{figure}

Figure~\ref{fig:weight_proj_clr_visual} visualizes the model weights in the approaches we analyzed for a 4-block ConvNet on the corruption, label flip, and rotation shift sequence. The oracle is denoted by a black X at the origin. The baseline that learns a single model using only data from the final time point is the point at the top of the plot. The model learned via ERM on only historical data is the starting point of the arrows for ERM plus fine-tuning and ERM plus linear probing then fine-tuning. For the sequential fine-tuning approaches and joint model, we plot the sequence of projections over all steps. A disadvantage of this approach is the side module weights are not the same dimension and cannot be visualized. We project the weights in each layer separately. The first 3 plots show the projections for the input convolution, the second block, and the output fully connected layer. Blocks 0, 2, and 3 behave similarly and are not shown. We also analyze the entire model by concatenating the weights from all layers and projecting this concatenated vector. This analysis is shown in the bottom right plot.

We hypothesized that the trajectory for each method from historical time points to the final distribution would head towards the oracle. While this was true for ERM followed by fine-tuning or linear probing then fine-tuning, the trajectories for all the other approaches are all congregated around the same point, making it impossible to distinguish which method leads itself to more optimal weights. This point does not get any closer in the $x$-direction to the oracle than the model fit with limited final sample size, so no method appears to derive any benefits from having access to historical data according to this visualization. Similarly, this point shares the same $y$-coordinate as a model trained only on historical data in the direction, so no method appears to benefit from having samples from the final distribution, again only according to this visualization. Based on the test accuracies, the approaches are able to perform better than a model trained without any historical data or a model trained without any data from the final time point. Thus, the collapse in the projection is more likely to be a limit of this visualization. In particular, the oracle is a rather arbitrary reference point for defining the origin and projecting all the models. If multiple models were given different random initializations and then trained on the same dataset of 20,000 samples from the final distribution, the weights would be very different, and the resulting plots may also appear different.

\textbf{Projection of intermediate outputs by canonical correlation analysis} While different random initializations can lead to wildly differently weights for both the oracle and the other methods we evaluate, we might expect that good-performing models would exhibit similar behavior. Thus, we decided to pivot towards comparing the activations from each layer instead. To do so, we use singular value canonical correlation analysis (SVCCA) to compare the outputs from intermediate layers in each model with an oracle model that was trained on a large number of samples from the final time step. Because we are interested in assessing whether the models align with the oracle at the final time point, we perform SVCCA on the intermediate activations when making predictions on the test set at the final time point. SVCCA was introduced by \citet{raghu2017svcca}. The first step of SVCCA is to obtain the singular values: The activations are standardized to have zero mean, and singular value decomposition is performed on these standardized activations to obtain the directions that explain the most variance. The second step is to compute the canonical correlation similarity. The subspaces are linearly transformed to produce pairs of directions with maximal correlation. The coefficients computed by SVCCA are the correlations between each pair of these transformed directions. \citet{raghu2017svcca} suggest summarizing the correlation by taking the mean of the top coefficients. We visualize the mean of the CCA coefficients explaining 50\% of the variance on the $x$-axis and the mean explaining 90\% of the variance on the $y$-axis. 

\begin{figure}[htbp]
    \centering
    \begin{minipage}[t]{.96\textwidth}
        \centering
        \includegraphics[width=\textwidth,trim={60 0 60 15},clip]{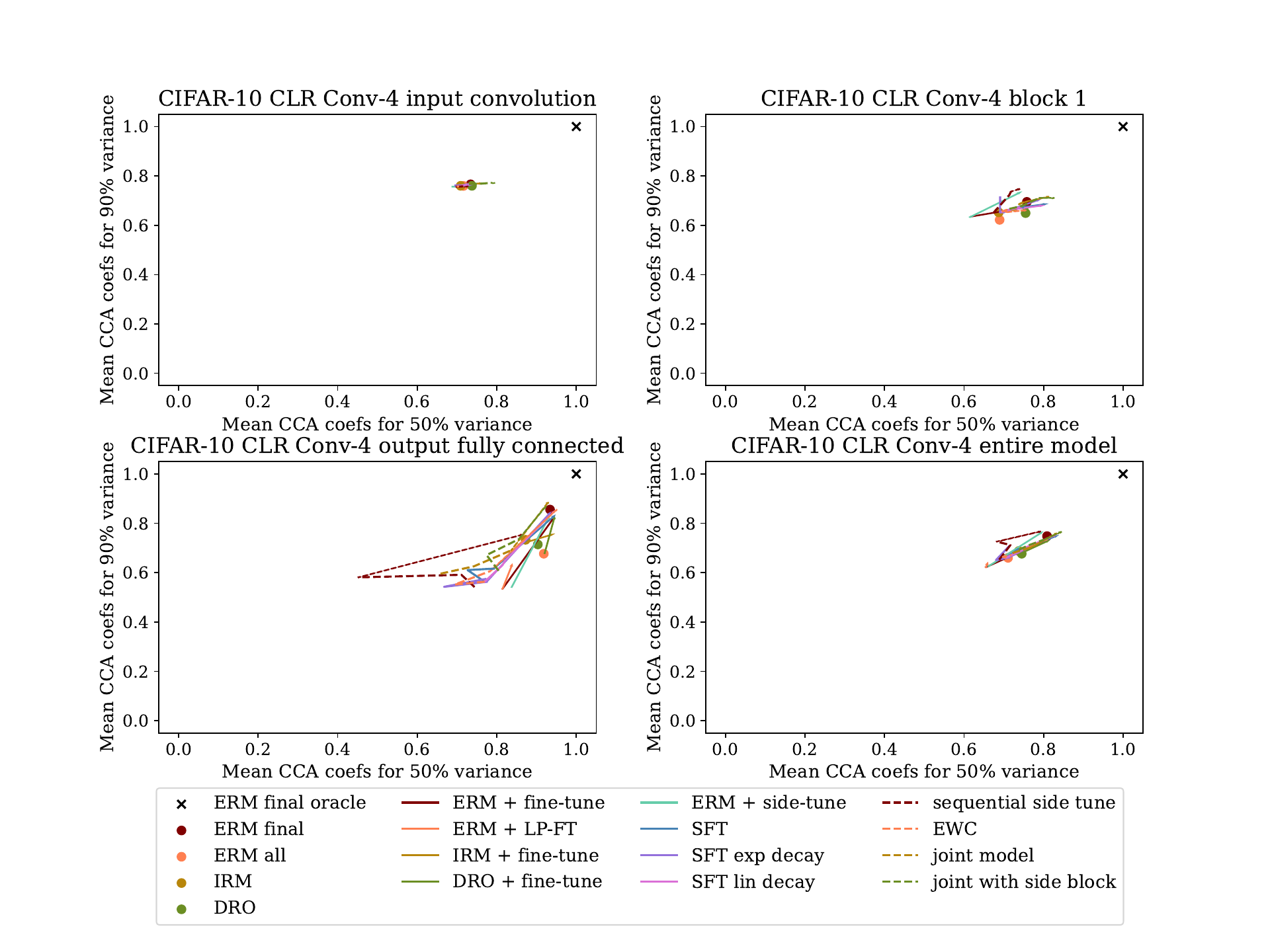}
    \end{minipage}
    \caption[Visualization of Conv-4 models over time for the corruption, label, and rotation shift sequence created by comparing the outputs from each layer to the oracle outputs via canonical correlation analysis.]{Visualization of Conv-4 models over time for the corruption, label, and rotation (CLR) shift sequence created by comparing the outputs from each layer to the oracle outputs via canonical correlation analysis (CCA). $x$-axis is mean of CCA coefficients explaining 50\% of variance when comparing model to final oracle. $y$-axis is mean explaining 90\%. Models closer to $\left(1, 1\right)$ have outputs that are more similar to the oracle. The subplot for the entire model concatenates outputs from all layers before running CCA.}
    \label{fig:cca_clr_visual}
\end{figure}

Figure~\ref{fig:cca_clr_visual} shows this visualization for the activations from the input convolutional layer, second block, and output fully connected layer in a 4-block ConvNet on the corruption, label flip, and rotation shift sequence. The activations from the output layer are taken before softmax is performed. We also analyze the entire model by concatenating the activations from all layers and performing SVCCA on this concatenated vector. This analysis is shown in the bottom right plot. These plots are slightly more interesting because the trajectories for different approaches do show changes over the time steps in the sequence, particularly for the fully connected layer. Another advantage of focusing on the activations is we can also visualize the side-tuning variants of each approach since we are not comparing each dimension of the parameter vector. The trajectories do show some progression towards the oracle at $\left(1, 1 \right)$. However, the methods still cannot be distinguished clearly, as many of them end up near the coordinate for performing ERM only on the limited final data. It seems like this visualization is also limited by the oracle being too arbitrary as a reference point.

To conclude, this appendix explores two ways to visualize how models change over time when adapting to a sequence of distribution shifts. The first visualization projects weights onto directions that were designed to reflect the shift from historical distributions and the limited sample size at the final time point. The second visualization focuses on the intermediate outputs instead since multiple parametrizations of the weights might lead to similar behavior. In both of these cases, the visualizations did not yield any insights because they were too dependent on an arbitrary reference point from a single training run of the oracle model. We hope this exploration will inspire the creation of more visualizations of models over time that are not dependent on an arbitrary reference point and yield fruitful insights into which approaches are better able to guide the model towards an optimum for the final time point.

\end{document}